\documentclass[sigconf]{acmart}
\setcopyright{none}
\acmConference[WWW '27]{The ACM Web Conference 2027}{May 10--14, 2027}{Dublin, Ireland}
\acmBooktitle{}
\acmYear{2027}
\copyrightyear{}
\acmDOI{}
\acmISBN{}

\usepackage{amsmath}
\usepackage{multirow}
\usepackage{array}
\usepackage{colortbl}
\usepackage{enumitem}
\usepackage{tcolorbox}
\usepackage{makecell}
\usepackage{needspace}

\tcbuselibrary{skins}
\definecolor{frameblue}{HTML}{315D91}
\definecolor{bgblue}{HTML}{F0F5FC}
\newtcolorbox{StrategyBox}[3][frameblue]{%
  enhanced,
  colframe=#1,
  colback=#2,
  colbacktitle=#1,
  coltitle=white,
  title={#3},
  fonttitle=\bfseries,
  fontupper=\small,
  boxrule=0.6pt,
  arc=2pt,
  left=5pt,right=5pt,top=4pt,bottom=4pt,
  before skip=8pt,after skip=8pt
}

\newcounter{idx}

\begin{document}

\title[AnomaMind]{AnomaMind: Coarse-to-Fine Agentic Diagnosis via Hybrid Inference for Time Series Anomaly Detection}

\author{Xiaoyu Tao}
\email{txytiny@mail.ustc.edu.cn}
\affiliation{%
  \institution{State Key Laboratory of Cognitive Intelligence, University of Science and Technology of China}
  \city{Hefei}
  \state{Anhui}
  \country{China}
}

\author{Yuchong Wu}
\email{wu3091455397@mail.ustc.edu.cn}
\affiliation{%
  \institution{State Key Laboratory of Cognitive Intelligence, University of Science and Technology of China}
  \city{Hefei}
  \state{Anhui}
  \country{China}
}

\author{Mingyue Cheng}
\email{mycheng@ustc.edu.cn}
\affiliation{%
  \institution{State Key Laboratory of Cognitive Intelligence, University of Science and Technology of China}
  \city{Hefei}
  \state{Anhui}
  \country{China}
}

\author{Ze Guo}
\email{gz1504921411@mail.ustc.edu.cn}
\affiliation{%
  \institution{State Key Laboratory of Cognitive Intelligence, University of Science and Technology of China}
  \city{Hefei}
  \state{Anhui}
  \country{China}
}

\author{Min Wu}
\affiliation{%
  \institution{Institute for Infocomm Research, Agency for Science, Technology and Research (A*STAR)}
  \city{Singapore}
  \country{Singapore}
}
\email{wumin@a-star.edu.sg}

\renewcommand{\shortauthors}{Tao et al.}

\begin{abstract}
Time series anomaly detection (TSAD) supports the analysis of web service key performance indicators (KPIs) by identifying anomalous intervals. Recent large language model (LLM)-based agentic approaches have achieved promising results through reasoning and tool use. Despite their effectiveness, precise anomaly localization remains challenging, as diverse anomaly patterns demand targeted numerical evidence and iterative refinement to accurately determine anomaly boundaries. In this work, we propose AnomaMind, a coarse-to-fine agentic diagnosis framework for TSAD. AnomaMind first identifies suspicious regions from time-series visualizations, then adaptively acquires targeted numerical evidence through multi-turn tool interactions, and finally refines anomaly decisions through self-reflection. This process is supported by a tool engine that integrates knowledge memory and numerical analysis.  To coordinate this diagnostic process, we design a hybrid inference strategy that employs frozen general-purpose LLMs for initial localization, tool orchestration, and evaluation, alongside a specialized LLM detector for evidence-grounded interval prediction. We optimize only the specialized detector through supervised fine-tuning (SFT) on diagnostic trajectories and reinforcement learning (RL) with verifiable detection rewards. Experiments on four benchmarks spanning web traffic, service KPIs, and system-monitoring signals, under both in-domain and multi-domain training settings, demonstrate competitive  performance and support the effectiveness of coarse-to-fine adaptive diagnosis through targeted evidence acquisition and self-reflective refinement.\footnote{The code is available at \url{https://github.com/Xiaoyu-Tao/AnomaMind-TS}.}

\end{abstract}


\maketitle

\section{Introduction}
Analyzing web service key performance indicators (KPIs), such as traffic volume, response latency, and error rates, helps characterize service behavior~\cite{ren2019time,zhang2022efficient}. These signals reflect both normal workload variation and potential service degradation~\cite{laptev2015generic,liu2015opprentice}. Time series anomaly detection (TSAD) identifies anomalous observations and delimits their temporal extent: overly broad intervals include normal behavior, while incomplete coverage misses part of an anomalous episode. We study anomaly interval localization in web service KPIs, asking how adaptive numerical inspection can support precise interval decisions across diverse KPI patterns.

Traditional TSAD methods rely on statistical or density-based criteria~\cite{schmidl2022anomaly,thill2017time,breunig2000lof}, while deep learning methods model temporal patterns through reconstruction or forecasting~\cite{xu2018unsupervised,audibert2020usad,munir2018deepant,deng2021graph}, with further advances in temporal association modeling~\cite{xuanomaly} and test-time adaptation~\cite{kim2024model}. Recent LLM-based approaches introduce contextual reasoning into anomaly detection. LLMAD combines retrieved examples with chain-of-thought prompting~\cite{liu2025large}, while AnomSeer employs reinforcement learning to improve multimodal anomaly reasoning and localization~\cite{zhang2026anomseer}. Beyond model-level reasoning, agentic approaches enable tool-guided detection and diagnosis. ARGOS generates executable detection rules through agent collaboration~\cite{gu2025argos}, TSAD-Agents integrates traditional detectors with multimodal LLMs for textual and visual anomaly analysis~\cite{xu2026multimodal}, and SAGE organizes specialized analyzers in a predefined parallel workflow for numerical evidence aggregation and diagnostic reporting~\cite{kang2026sage}. Despite these advances, precise anomaly localization remains challenging, as different anomaly patterns require targeted numerical evidence beyond initial observations. This motivates a coarse-to-fine diagnostic process that adaptively acquires evidence and refines anomaly intervals based on intermediate findings.

Addressing this research question presents two key challenges. First, refining coarse anomaly hypotheses requires acquiring relevant numerical evidence according to the suspected anomaly patterns and intermediate findings. Value deviations, temporal changes, and regional observations provide complementary clues, whose relevance depends on the suspected behavior. For example, interpreting a drop in web traffic requires examining its magnitude, transition strength, and surrounding observations against the expected temporal pattern. A predefined diagnostic path may overlook useful evidence, while intermediate findings may reveal the need for further investigation. The diagnostic process must therefore adapt subsequent tool interactions to the available evidence and revisit uncertain regions when necessary~\cite{he2026harnessing}. Second, the acquired evidence must be interpreted to determine precise anomaly boundaries. A sharp transition may indicate suspicious behavior, but its presence alone does not determine the temporal extent of an anomaly. General-purpose LLMs may struggle to translate heterogeneous diagnostic findings into accurate interval predictions~\cite{alnegheimish2024large,parkdelving}, motivating task-specific detector learning to support precise localization within the adaptive diagnostic process.

Our key insight is to organize anomaly localization as a coarse-to-fine diagnostic process driven by evolving evidence and iterative feedback. General-purpose LLMs first identify suspicious regions from time-series visualizations and then coordinate contextual knowledge and numerical analysis tools to investigate these regions~\cite{yao2023react}. Subsequent tool interactions are guided by intermediate diagnostic findings, allowing the process to acquire targeted evidence as the diagnostic context evolves. A specialized detector interprets the accumulated evidence to predict anomaly intervals, while an evaluator reviews these predictions and can request further investigation when inconsistencies arise~\cite{shinn2023reflexion}. To support precise localization, we optimize the specialized detector through supervised fine-tuning on diagnostic trajectories and reinforcement learning with verifiable detection rewards, while keeping the general-purpose LLMs frozen. This design combines flexible evidence-guided diagnosis with task-specific interval learning, enabling iterative refinement of anomaly decisions without training the entire diagnostic workflow.

Guided by this insight, we propose AnomaMind, a coarse-to-fine agentic diagnosis framework for TSAD. AnomaMind first identifies suspicious regions from time-series visualizations, then adaptively acquires numerical evidence through multi-turn tool interactions, and finally refines anomaly decisions through self-reflection. To support this process, we develop a reusable diagnostic tool engine that integrates contextual knowledge and numerical analysis. Knowledge memory provides contextual guidance, while numerical tools offer targeted evidence for anomaly assessment. Intermediate findings guide subsequent tool interactions, allowing the diagnostic process to adapt as new evidence emerges. A specialized LLM detector predicts anomaly intervals from the accumulated evidence, while an evaluator reviews these predictions and can request further diagnosis and refinement. To coordinate this process, AnomaMind adopts a hybrid inference strategy that employs frozen general-purpose LLMs for initial localization, tool orchestration, and evaluation, alongside a specialized detector for interval prediction. We optimize only the detector through SFT on diagnostic trajectories and RL with verifiable detection rewards.

Our main contributions are summarized as follows:
\begin{itemize}

\item We propose AnomaMind, a coarse-to-fine agentic diagnosis framework for TSAD that combines initial localization, targeted evidence acquisition guided by intermediate findings, and self-reflective refinement.

\item We develop a hybrid inference strategy that keeps general-purpose LLMs frozen for diagnostic reasoning while optimizing a specialized detector through SFT on diagnostic trajectories and RL with verifiable detection rewards.
    
    \item We conduct extensive experiments on four benchmarks to evaluate detection performance, diagnostic workflow components, detector optimization, and inference efficiency, revealing dataset-dependent effects and trade-offs between detection quality and computational cost.
\end{itemize}

\section{Related Work}

\subsection{Time Series Anomaly Detection}
Existing TSAD methods can be broadly categorized into statistical and density-based approaches, deep learning-based approaches, and LLM-based approaches~\cite{deforce2025tsap}. Traditional methods identify anomalies through statistical deviations or density-based criteria~\cite{chandola2009anomaly,zhang2017cleaning}. Deep learning methods primarily detect anomalies through reconstruction or forecasting errors~\cite{su2019robust,audibert2020usad,munir2018deepant,zhang2019mscred}. Beyond these approaches, recent studies explore temporal dependencies and multi-resolution representations. Anomaly Transformer models association discrepancies~\cite{xuanomaly}, while NCAD learns contextual representations using synthetic anomalies~\cite{carmona2022neural}. FCVAE incorporates global and local frequency features~\cite{wang2024fcvae}, Dual-TF aligns time- and frequency-domain anomaly scores~\cite{nam2024dualtf}, and HYDRA aggregates multi-resolution anomaly evidence~\cite{huang2026hydra}.
LLM-based approaches extend TSAD through prompting, representation learning, and anomaly reasoning~\cite{wu2026icad}. SigLLM explores direct prompting and forecasting-based detection~\cite{alnegheimish2024large}, while AnomalyLLM identifies teacher--student representation discrepancies~\cite{liu2024large}. LLMAD combines retrieved normal and anomalous examples with anomaly-specific chain-of-thought prompting without task-specific fine-tuning~\cite{liu2025large}. AnomSeer studies model-level anomaly reasoning, using ExpCoT-guided reinforcement learning with TimerPO to improve a single multimodal LLM's classification, localization, and explanation~\cite{zhang2026anomseer}. Other studies improve numerical interpretation through statistical decomposition and index-aware prompting~\cite{parkdelving}, while Time-RA introduces an anomaly-diagnosis task and a feedback-curated dataset~\cite{yang2026timera}. These studies develop models' interpretation and prediction capabilities; agentic diagnosis additionally concerns how evidence is acquired and revisited during inference.

\subsection{LLM-Based Agents}

LLM-based agents combine multi-step reasoning, external tool interaction, and iterative feedback~\cite{zhang2026search}. ReAct interleaves reasoning with information-seeking actions~\cite{yao2023react}, while Reflexion improves decisions through linguistic feedback and episodic memory without updating model weights~\cite{shinn2023reflexion}. Learning-based approaches include Toolformer, which learns API invocation and result integration through self-supervision~\cite{schick2023toolformer}, and ToolRL, which optimizes tool selection and execution through reinforcement learning~\cite{qian2025toolrl}.
For time series, a recent survey organizes LLM-based reasoning and agentic systems by reasoning topology and task objective~\cite{chang2026survey}. TimeOmni-1 improves temporal reasoning through SFT and RL~\cite{guan2026timeomni}. ARGOS generates and validates executable detection rules whose execution requires no LLM calls~\cite{gu2025argos}. TSAD-Agents routes point anomalies to traditional detectors and range anomalies to visual MLLMs, then refines predictions~\cite{xu2026multimodal}. SAGE organizes four specialized analyzers in a predefined parallel workflow, followed by evidence aggregation and diagnostic reporting~\cite{kang2026sage}. AnomaMind focuses on adaptive agentic diagnosis: intermediate findings guide subsequent tool queries, and review feedback can initiate further evidence acquisition. A specialized detector learns to map the acquired evidence to anomaly intervals through SFT and RL.

\begin{figure*}[t]
    \centering
    \includegraphics[width=\linewidth]{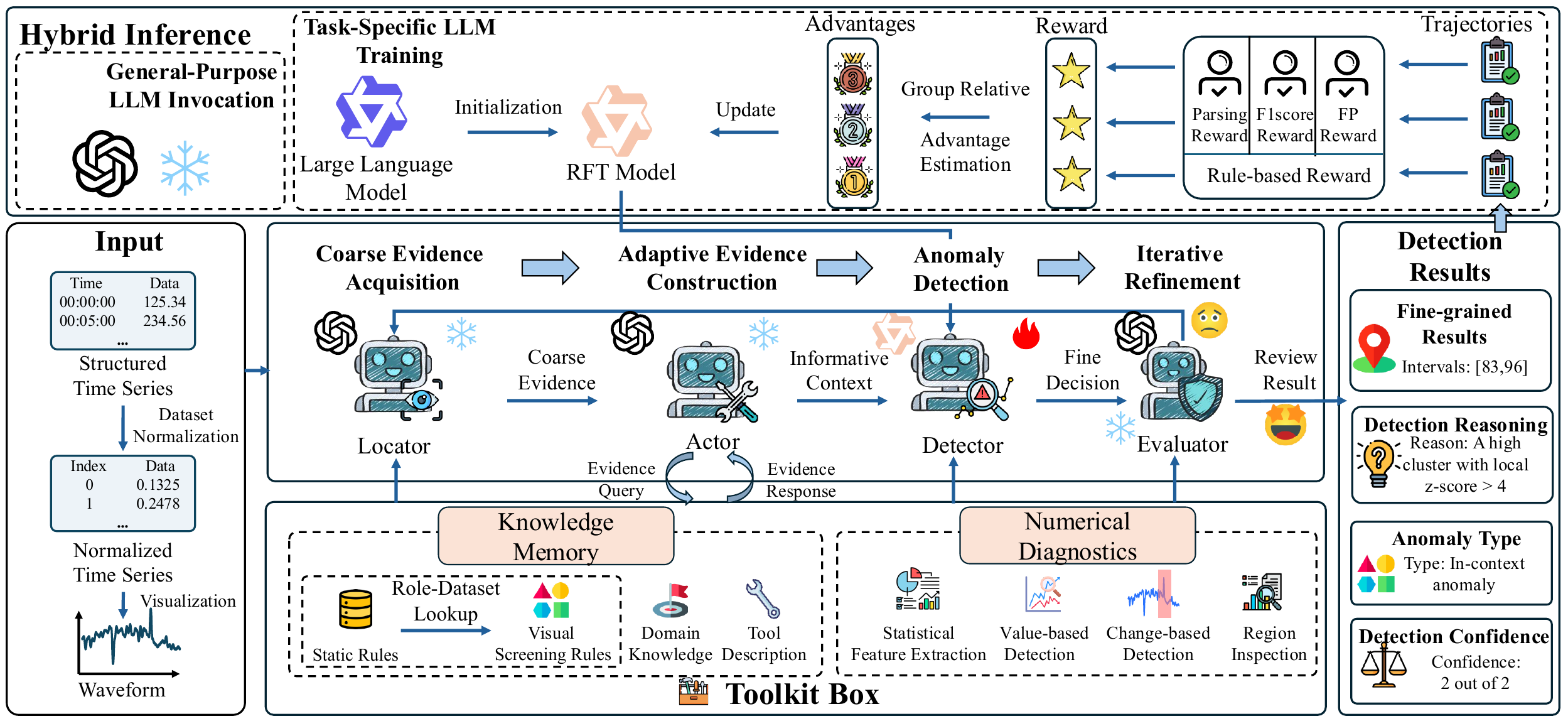}
    \caption{Overview of AnomaMind. A coarse-to-fine workflow integrates adaptive evidence acquisition through contextual knowledge and numerical tools with self-reflective interval refinement. Hybrid inference keeps general-purpose LLMs frozen while optimizing the specialized detector through SFT on diagnostic trajectories and RL with verifiable detection rewards.}
    \Description{The Locator proposes suspicious intervals, the Actor obtains numerical evidence, the Detector predicts anomaly intervals, and the Evaluator reviews the result and may request another diagnostic pass.}
    \label{fig:framework}
\end{figure*}

\section{Preliminaries}

\subsection{Problem Formulation}
The goal of interval-level TSAD is to identify anomalous regions in a fully observed input window:
\begin{equation}
\hat{S}=\{[b_j,e_j]\}_{j=1}^{M},
\qquad 1\leq b_j\leq e_j\leq T,
\end{equation}
where $\hat{S}$ is the predicted interval set for the input window $\mathbf x_{1:T}=\{x_1,\ldots,x_T\}$ of length $T$, $b_j$ and $e_j$ are the inclusive start and end boundaries of the $j$-th interval, respectively, and $M$ is the number of detected intervals. An empty set denotes no detected anomaly. The corresponding point-wise prediction is:
\begin{equation}
\hat y_t=\mathrm{I}\left\{t\in\bigcup_{j=1}^{M}[b_j,e_j]\right\},
\end{equation}
where $\hat y_t$ is the binary anomaly prediction at observation $t$, and $\mathrm{I}\{\cdot\}$ is the indicator function, equal to 1 when $t$ lies in at least one predicted interval and 0 otherwise. This representation covers both isolated anomalous points and contiguous abnormal segments. Detection requires determining whether suspicious behavior constitutes an anomaly and locating its temporal extent. Explanations and ordinal confidence accompany the intervals to support interpretation, while interval coverage defines the detection prediction.

\subsection{Tool-augmented Detection}

Let $t$ index observations in a fixed window and $k$ index tool interactions within a diagnostic pass. Candidate regions provide coarse hypotheses that may differ from the final anomaly intervals. At step $k$, the diagnostic state $h_k$ contains the observed window, contextual guidance, candidates, previous tool interactions, and review feedback. An action $a_k$ selects a tool and query to acquire evidence:
\begin{equation}
\begin{aligned}
o_k &= F_{a_k}(\mathbf{x}, B),\\
h_{k+1} &= U(h_k, a_k, o_k),
\end{aligned}
\end{equation}
where $B$ denotes numerical reference information, $o_k$ is the returned evidence, and $U$ updates the diagnostic state. The resulting context guides anomaly interval prediction.

This process integrates three capabilities. Adaptive evidence acquisition selects information based on intermediate findings. Evidence-grounded interval prediction interprets the acquired evidence and temporal context to predict anomaly intervals. Self-reflective refinement reviews predictions for inconsistencies and initiates further diagnosis when needed. Evidence accumulates within each pass, and inference ends upon acceptance or reaching the interaction budget. Evaluation uses no test labels, while training uses verifiable rewards to optimize the detector's evidence-to-interval mapping. General-purpose LLMs remain frozen, with contexts held fixed when comparing alternative predictions.

\section{The Proposed AnomaMind}
\label{sec:method}

\subsection{Framework Overview}
\label{sec:framework-overview}
As shown in Figure~\ref{fig:framework}, AnomaMind implements adaptive agentic diagnosis through a coarse-to-fine workflow of adaptive evidence acquisition, interval prediction, and self-reflective refinement. The locator first proposes suspicious intervals, and the actor gathers targeted numerical evidence through multi-turn tool interactions. The detector then synthesizes this evidence into fine-grained anomaly intervals, while the evaluator reviews the diagnosis and can request another diagnostic pass. AnomaMind adopts a hybrid inference strategy: frozen general-purpose LLMs handle initial localization, tool orchestration, and evaluation, while only the detector policy $\pi_\theta$ is trained. Supervised fine-tuning (SFT) initializes $\pi_\theta$, and reinforcement learning (RL) optimizes its anomaly decisions using verifiable detection rewards.

\subsection{Adaptive Evidence Acquisition with a Reusable Tool Engine}
\label{sec:adaptive-feature-preparation}
The locator proposes candidate intervals by combining visual evidence with role-specific contextual guidance:
\begin{equation}
C=L_{\mathrm{loc}}\bigl(V(\mathbf x),K_{loc}\bigr),
\end{equation}
where $C$ denotes candidate intervals, $L_{\mathrm{loc}}$ is the locator, $V(\mathbf x)$ is the visual representation of input window $\mathbf x$, and $K_{loc}$ provides role-specific contextual guidance. Candidates identify behavior meriting inspection and provide an interpretation. Their boundaries are provisional: the detector uses numerical evidence and contextual guidance to assess anomaly presence and refine boundaries. Localization supplies hypotheses for diagnostic analysis.

To examine these hypotheses, AnomaMind's tool engine integrates knowledge memory $K$ with statistical, value-based, change-based, and regional numerical tools, denoted by $N=\{N_{stat},N_{val},N_{chg},N_{reg}\}$. Statistical summaries characterize the current window; value-based measurements reveal departures from reference levels; change-based measurements describe temporal transitions; regional inspection returns observations at queried positions. These views are complementary. A persistent deviation may occupy an unusual value range while exhibiting modest internal changes, whereas a brief disturbance may have pronounced transitions at its boundaries. Knowledge memory $K$ supplies role-specific domain expectations and interpretation rules for these measurements. A tool returns evidence as $o=F_a(\mathbf x,B)$, where $a$ specifies a tool operator and its query, and $B$ denotes its numerical reference information. The response includes measurements and their locations. The reference construction and preprocessing scope are specified in Appendix~\ref{app:knowledge-tools} and Appendix~\ref{app:data-preprocessing}.

Using these resources, the actor selects diagnostics according to the candidates, contextual guidance, and evidence already obtained. Within each diagnostic pass, acquisition proceeds as follows:
\begin{equation}
\begin{aligned}
a_{r,k}&=A_{\mathrm{act}}\bigl(C,Z^{(r)}_{<k},
K_{act},f^{(r)}\bigr),\\
o_{r,k}&=F_{a_{r,k}}(\mathbf x,B),\\
Z^{(r)}_{1:k}&=Z^{(r)}_{<k}\mathbin{\|}(a_{r,k},o_{r,k}),
\end{aligned}
\end{equation}
where $r$ and $k$ index diagnostic passes and tool actions, respectively; $A_{\mathrm{act}}$ denotes the actor with contextual guidance $K_{act}$; and $a_{r,k}$ and $o_{r,k}$ are the selected query and returned evidence. The ordered transcripts $Z^{(r)}_{<k}$ and $Z^{(r)}_{1:k}$ record interactions before and through action $k$, respectively, with $\|$ appending a query--result pair. Feedback $f^{(r)}$ contains the latest review and is initially empty. Intermediate findings can thus guide later tool choices, score thresholds, and regional queries. For example, a sharp transition can motivate inspecting nearby observations to assess its duration. Adaptive evidence acquisition consists of selecting and organizing these diagnostic views. The transcript preserves both the measurements and the queries that produced them, forming an evidence base for the detector. A tool response contributes to the diagnosis without independently assigning the final anomaly label.

\subsection{Evidence-Grounded Interval Prediction}
\label{sec:reasoning-aware-detection}
The detector synthesizes the candidate hypotheses, tool evidence, contextual expectations, and any review feedback. Their assembled diagnostic context is:
\begin{equation}
\mathbf d^{(r)}=\Psi\bigl(C,Z^{(r)},
K_{det},f^{(r)}\bigr),
\end{equation}
where $\mathbf d^{(r)}$ is the diagnostic context for pass $r$, and $\Psi$ organizes candidate intervals $C$, the completed interaction transcript $Z^{(r)}$, detector-specific guidance $K_{det}$, and the latest review feedback $f^{(r)}$ for the detection policy. Candidates direct attention, query--result pairs supply numerical observations, and contextual guidance supports their interpretation.

Evidence-grounded interval prediction connects these sources to two judgments: whether the observed behavior is anomalous in context, and which observations belong to the anomaly. An unusual magnitude can be interpreted alongside local temporal structure and expected domain behavior. Evidence of an abrupt change can help assess a candidate boundary, while regional observations show whether the departure persists or returns to the surrounding pattern. The policy can consequently revise the initial visual interpretation when numerical evidence supports another conclusion. The detector expresses these judgments through a structured response:
\begin{equation}
u^{(r)}\sim\pi_\theta(\,\cdot\mid\mathbf d^{(r)}),
\qquad \hat{S}^{(r)}=\operatorname{Parse}(u^{(r)}),
\end{equation}
where $\pi_\theta$ is the detector policy, $u^{(r)}$ is its response containing an explanation and structured anomaly intervals with ordinal confidence, and $\operatorname{Parse}$ extracts the predicted interval set $\hat{S}^{(r)}$. An empty interval set indicates that no anomaly is detected. The detector provides explanations that relate its predictions to the acquired diagnostic evidence, while predicted intervals define the final detection results. Explanations and ordinal confidence scores serve as auxiliary outputs to support the interpretation of anomaly decisions. Appendix~\ref{app:workflow-control} details the output format and parsing rules.

\subsection{Self-Reflective Refinement}
\label{sec:self-reflective-refinement}
A provisional decision may conflict with the acquired evidence or leave an important boundary insufficiently supported. The evaluator reviews summaries of the actor's evidence and the detector's response, together with prior feedback when available. Its task is to identify serious inconsistencies and provide a suggestion for further diagnosis. This review uses the supplied diagnostic context without access to test labels.

If the result is accepted, the workflow terminates. Otherwise, the feedback $f^{(r+1)}$ guides a new actor--detector pass. The original candidates $C$ are retained, while a fresh transcript is constructed through renewed tool interaction. The actor can obtain evidence relevant to the identified issue, and the detector produces another judgment from the rebuilt context. For example, uncertainty about an interval endpoint can motivate examining observations around that boundary before making the next decision. Evidence accumulates within a pass; cross-pass guidance is carried by the latest review feedback. A finite pass budget bounds the process, with termination upon acceptance or budget exhaustion and explicit handling of invalid final outputs. The review interface, restart semantics, and stopping rules are specified in Appendix~\ref{app:workflow-control}. Inference-time review provides evidence-based consistency feedback; the label-based learning signal below optimizes the detector during training.

\subsection{Detection-Specific Policy Optimization}
\label{sec:detection-policy-optimization}
SFT initializes the detector from supervised context--response pairs. The training objective is:
\begin{equation}
L_{SFT}(\theta)=
-\frac{\sum_i\log\pi_\theta(u_i^*\mid\mathbf d_i)}
{\sum_i|u_i^*|},
\end{equation}
where $\theta$ denotes the detector parameters, $D_b=\{(\mathbf d_i,u_i^*)\}$ is a batch from the SFT dataset $D_{SFT}$, and each pair consists of a diagnostic context $\mathbf d_i$ and its target response $u_i^*$. The length $|u_i^*|$ counts retained supervised response tokens, excluding prompt and padding. This teaches the policy to interpret diagnostic inputs and produce structured decisions. Data construction and token-level supervision are detailed in Appendix~\ref{app:training-inference}.

Building on this initialization, RL aligns the policy with the detection objective. The reward evaluates the structured prediction. On anomalous windows, its task component measures point-wise F1 between the predicted interval coverage and labeled anomalous points. On normal windows, it decreases with false-positive coverage and is maximized by an empty prediction. Format checks make the output evaluable. The expected-reward objective is:
\begin{equation}
J(\theta)=\mathrm{E}_{(\mathbf d,\mathbf y)\sim D_{RL}}
\mathrm{E}_{u\sim\pi_\theta(\cdot\mid\mathbf d)}
\bigl[R(u,\mathbf y)\bigr],
\end{equation}
where $D_{RL}$ contains diagnostic contexts $\mathbf d$ paired with point-wise training labels $\mathbf y$, $u$ is a response sampled from the detector policy $\pi_\theta$, and $R(u,\mathbf y)$ is the reward assigned to its structured prediction. The reward assesses completed detector outputs against training labels. It supplies no separate score for explanation quality or tool-selection trajectories; explanation faithfulness is not separately evaluated. Appendix~\ref{sec:detector-reward} specifies the reward and parsing rules.

Group-relative policy optimization samples multiple responses and compares their rewards. The normalized advantage is:
\begin{equation}
\hat A_i=\frac{r_i-\bar r}{s_r+\varepsilon},
\end{equation}
where $r_i$ is the reward of response $u_i$, $\bar r$ and $s_r$ are the group's reward mean and sample standard deviation, and $\varepsilon$ stabilizes normalization. A clipped policy surrogate uses these advantages to favor better-performing responses relative to their peers. The token-level objective is provided in Appendix~\ref{sec:grpo-updates}. All responses within a group share the same diagnostic context, so their comparison holds the acquired evidence fixed. Both SFT and RL update only $\pi_\theta$; the locator, actor, and evaluator remain frozen and receive no policy-gradient credit. This learning scheme places task-specific optimization of the evidence-to-decision mapping.

\begin{table}[t]
\centering
\small
\caption{Statistics of the four TSAD benchmarks.}
\label{tab:dataset}
\resizebox{1\columnwidth}{!}{%
\begin{tabular}{c|cccc}
\toprule
Datasets & \multicolumn{1}{r}{Total Points} &
  \multicolumn{1}{r}{Train Points} &
  \multicolumn{1}{r}{Test Points} &
  \multicolumn{1}{r}{Anomaly Type}  \\ \midrule
YAHOO & 182,600 & 107,200 & 75,400 & {Point \& Segment} \\
KPI   & 1,071,000 & 483,000 & 588,000 & {Segment} \\
IOPS  & 309,000 & 147,000 & 162,000 & {Segment} \\
WSD   & 203,000 & 89,000 & 114,000 & {Segment} \\\bottomrule
\end{tabular}%
}
\end{table}

\begin{table*}[t]
\centering
\small
\caption{Comparison of anomaly detection performance across multiple benchmark datasets. The best and second-best results are highlighted in \textbf{bold} and \underline{underlined}, respectively.}
\label{tab:main-new}
\footnotesize
\setlength{\tabcolsep}{4.0pt}
\resizebox{\linewidth}{!}{%
\begin{tabular}{c| *{16}{c}}
\toprule
\multirow{2}{*}{Methods}
& \multicolumn{4}{c}{YAHOO}
& \multicolumn{4}{c}{KPI}
& \multicolumn{4}{c}{WSD}
& \multicolumn{4}{c}{IOPS} \\
\cmidrule(lr){2-5} \cmidrule(lr){6-9} \cmidrule(lr){10-13} \cmidrule(lr){14-17}
& F1 & Best-F1 & AUC-PR & Range-F1
& F1 & Best-F1 & AUC-PR & Range-F1
& F1 & Best-F1 & AUC-PR & Range-F1
& F1 & Best-F1 & AUC-PR & Range-F1 \\
\midrule
DWT\_MLEAD
& 0.268 & 0.314 & 0.205 & 0.407 & 0.227 & 0.306 & 0.181 & 0.360 & 0.472 & 0.614 & 0.534 & 0.642 & 0.388 & 0.451 & 0.413 & 0.431 \\
FFT
& 0.361 & 0.689 & 0.623 & 0.695 & 0.267 & 0.305 & 0.194 & 0.373 & 0.159 & 0.230 & 0.186 & 0.292 & 0.172 & 0.287 & 0.181 & 0.389 \\
SR
& 0.573 & 0.727 & 0.657 & 0.759 & 0.293 & 0.316 & 0.273 & 0.573 & 0.296 & 0.539 & 0.461 & 0.625 & 0.127 & 0.419 & 0.290 & 0.542 \\
MatrixProfile
& 0.459 & 0.641 & 0.542 & 0.658 & 0.396 & 0.407 & 0.364 & 0.529 & 0.324 & 0.468 & 0.347 & 0.460 & 0.369 & 0.453 & 0.381 & 0.486 \\
Sub\_PCA
& 0.249 & 0.364 & 0.219 & 0.385 & 0.219 & 0.475 & 0.376 & 0.557 & 0.381 & 0.475 & 0.416 & 0.589 & 0.453 & 0.523 & 0.492 & 0.538 \\
CNN
& 0.408 & 0.620 & 0.573 & 0.661 & 0.214 & 0.313 & 0.254 & 0.512 & 0.308 & 0.528 & 0.432 & 0.651 & 0.175 & 0.415 & 0.323 & 0.537 \\
M2N2
& 0.379 & 0.523 & 0.476 & 0.578 & 0.184 & 0.316 & 0.255 & 0.498 & 0.216 & 0.483 & 0.402 & 0.597 & 0.178 & 0.496 & 0.384 & 0.493 \\
LSTMAD
& 0.426 & 0.642 & 0.595 & 0.694 & 0.161 & 0.287 & 0.227 & 0.535 & 0.422 & 0.609 & 0.520 & 0.664 & 0.161 & 0.425 & 0.340 & 0.528 \\
TranAD
& 0.253 & 0.406 & 0.355 & 0.445 & 0.359 & 0.386 & 0.264 & 0.372 & 0.199 & 0.300 & 0.217 & 0.431 & 0.238 & 0.436 & 0.352 & 0.420 \\
OmniAnomaly
& 0.351 & 0.515 & 0.467 & 0.529 & 0.400 & 0.471 & \underline{0.417} & 0.389 & 0.226 & 0.334 & 0.298 & 0.448 & 0.265 & 0.386 & 0.315 & 0.439 \\
NCAD (sup.)
& 0.453 & 0.534 & 0.510 & 0.527 & 0.276 & 0.323 & 0.251 & 0.448 & 0.510 & 0.572 & 0.497 & 0.567 & 0.281 & 0.299 & 0.269 & 0.491 \\
Chronos
& 0.698 & 0.789 & 0.782 & 0.863 & 0.185 & 0.284 & 0.220 & 0.493 & 0.305 & 0.515 & 0.410 & 0.581 & 0.149 & 0.488 & 0.375 & 0.543 \\
TimesFM
& 0.681 & 0.849 & 0.826 & \underline{0.896} & 0.197 & 0.312 & 0.245 & 0.538 & 0.355 & 0.546 & 0.429 & 0.603 & 0.147 & 0.435 & 0.350 & 0.354 \\
OFA
& 0.491 & 0.723 & 0.684 & 0.753 & 0.400 & 0.407 & 0.375 & 0.415 & 0.294 & 0.464 & 0.384 & 0.500 & 0.551 & 0.567 & 0.450 & 0.652 \\
LLM-AD
& 0.352 & 0.371 & 0.220 & 0.362 & 0.381 & 0.417 & 0.327 & 0.538 & 0.293 & 0.294 & 0.193 & 0.372 & 0.277 & 0.354 & 0.256 & 0.533 \\
LLM-TSAD
& 0.626 & 0.626 & 0.553 & 0.669 & 0.437 & 0.437 & 0.258 & 0.527 & 0.422 & 0.422 & 0.289 & 0.470 & 0.316 & 0.316 & 0.153 & 0.465 \\
ARGOS
& 0.484 & 0.484 & 0.509 & 0.590 & 0.477 & 0.477 & 0.288 & 0.548 & 0.364 & 0.364 & 0.358 & 0.408 & 0.301 & 0.301 & 0.267 & 0.236 \\
TSAD-Agents
& 0.719 & 0.719 & 0.703 & 0.793
& 0.449 & 0.449 & 0.345 & 0.517
& 0.562 & 0.562 & 0.443 & 0.610
& 0.669 & 0.669 & 0.442 & 0.618 \\

AnomaMind-Direct 
& 0.526 & 0.526 & 0.497 & 0.542 & 0.367 & 0.367 & 0.287 & 0.453 & 0.392 & 0.392 & 0.288 & 0.442 & 0.283 & 0.283 & 0.247 & 0.472 \\
AnomaMind-Zero
& 0.848 & 0.861 & 0.814 & 0.859
& 0.548 & 0.571 & 0.341 & 0.752
& 0.646 & 0.650 & 0.529 & \underline{0.777}
& 0.761 & 0.766 & \underline{0.698} & 0.836 \\
\midrule
\rowcolor[gray]{0.95} \textbf{Ours (In-domain) }
& \textbf{0.902} & \textbf{0.909} & \textbf{0.880} & \textbf{0.903} & \textbf{0.679} & \textbf{0.682} & \textbf{0.519} & \textbf{0.804} & \underline{0.697} & \underline{0.702} & \underline{0.587} & 0.721 & \underline{0.801} & \underline{0.814} & 0.672 & \underline{0.840} \\
\rowcolor[gray]{0.95} \textbf{Ours(Multi-domain)}
& \underline{0.891} & \underline{0.896} & \underline{0.856} & 0.892 & \underline{0.606} & \underline{0.618} & 0.403 & \underline{0.800} & \textbf{0.805} & \textbf{0.816} & \textbf{0.730} & \textbf{0.842} & \textbf{0.855} & \textbf{0.865} & \textbf{0.767} & \textbf{0.909} \\
\bottomrule
\end{tabular}
}
\end{table*}

\section{Experiments}
\label{sec:experiments}

\subsection{Experimental Settings}
\label{sec:experimental-settings}

\paragraph{Datasets.}
Table~\ref{tab:dataset} summarizes benchmarks used to assess anomaly interval localization in web service KPI analysis; three are included in the recent TSB-AD benchmark~\cite{liu2024elephant}. YAHOO~\cite{laptev2015generic} contains real production traffic and synthetic time series. KPI and WSD~\cite{zhang2022efficient} provide performance indicators collected from large-scale web services. IOPS~\cite{xu2018unsupervised} includes service performance and machine health metrics. Test samples are constructed without anomaly-label-based selection, and all methods use identical test splits and the same preprocessing pipeline. Preprocessing and numerical references are computed separately over complete training and test splits; test anomaly labels are reserved for evaluation. Appendix~\ref{app:sample-construction} describes data splitting, windowing, and preprocessing.

\paragraph{Baselines.}
We compare AnomaMind with representative baselines across five categories. Statistical methods include DWT-MLEAD~\cite{thill2017time}, FFT-AD, MatrixProfile~\cite{yeh2016matrix}, SR~\cite{ren2019time}, and Sub-PCA~\cite{shyu2003novel}. Deep learning methods include CNN~\cite{munir2018deepant}, M2N2~\cite{kim2024model}, LSTMAD~\cite{malhotra2015long}, TranAD~\cite{tuli2022tranad}, OmniAnomaly~\cite{su2019robust}, and NCAD~\cite{carmona2022neural}. Pretrained models include TimesFM~\cite{das2024decoder}, Chronos~\cite{ansari2024chronos}, and OFA~\cite{zhou2023one}. LLM-based methods include LLM-AD~\cite{liu2025large} and LLM-TSAD~\cite{parkdelving}, while agentic methods include ARGOS~\cite{gu2025argos} and TSAD-Agents~\cite{xu2026multimodal}. We include two AnomaMind variants to examine the effects of diagnostic interactions and task-specific learning. AnomaMind-Zero retains the complete diagnostic workflow with frozen general-purpose LLMs, while AnomaMind-Direct uses SFT for direct anomaly prediction without diagnostic interactions. These variants provide complementary system-level comparisons under different inference and training configurations. Baselines use official implementations where available, and supervised NCAD and AnomaMind-Direct share AnomaMind's labeled training splits. Further details are provided in Appendix~\ref{app:baselines}.

\paragraph{Implementation.}
AnomaMind employs Qwen3-8B~\cite{yang2025qwen3} as the trainable detector and Gemini-3.1-flash for frozen general-purpose reasoning roles. The detector undergoes two-stage full-parameter training. SFT uses recorded diagnostic trajectories as supervision, with a learning rate of $5\times10^{-5}$ for one epoch. GRPO further optimizes detection performance using verifiable rewards derived from annotated anomaly intervals, with a learning rate of $10^{-6}$, a group size of 8, and two training epochs. Training uses eight Ascend NPUs. We evaluate both in-domain models trained separately on each dataset and a multi-domain model jointly trained across all four datasets. Results are averaged over three inference runs using the same trained checkpoint. Additional training and inference details are provided in Appendix~\ref{app:training-inference}.

\begin{figure}[!t]
\centering
\includegraphics[width=\linewidth]{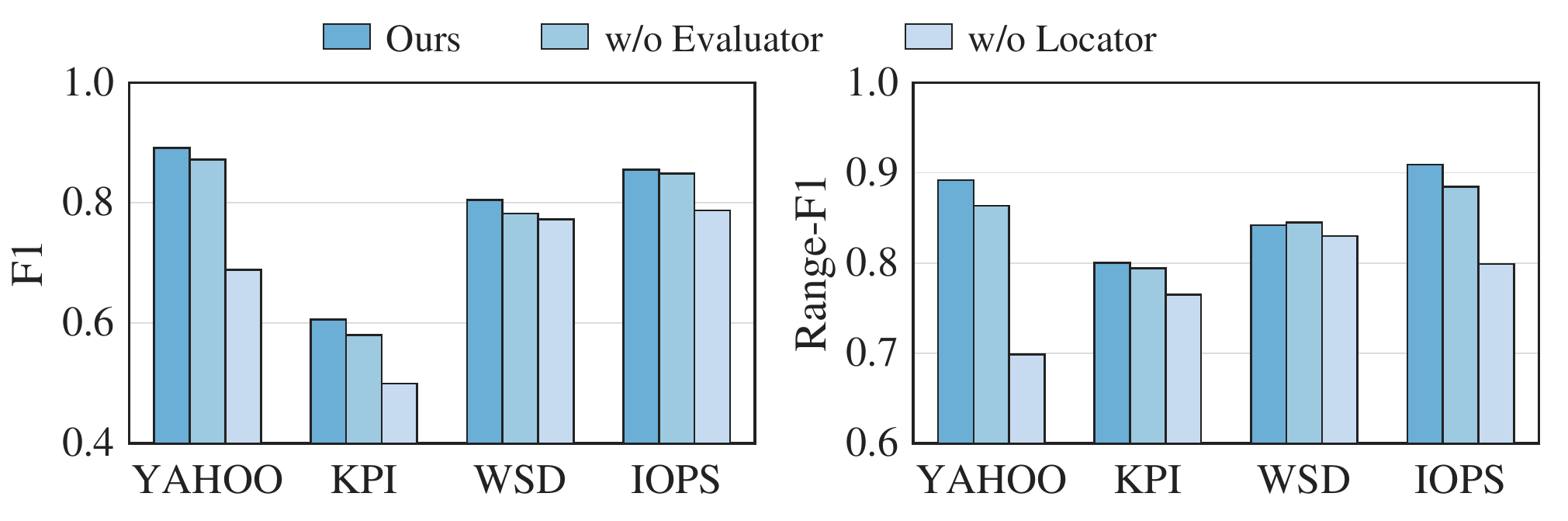}
\caption{Ablation study on the contributions of different roles in the diagnostic workflow. }
\Description{F1 and Range-F1 for the complete workflow and variants without the Evaluator or Locator across four datasets.}
\label{fig:workflow}

\includegraphics[width=\linewidth]{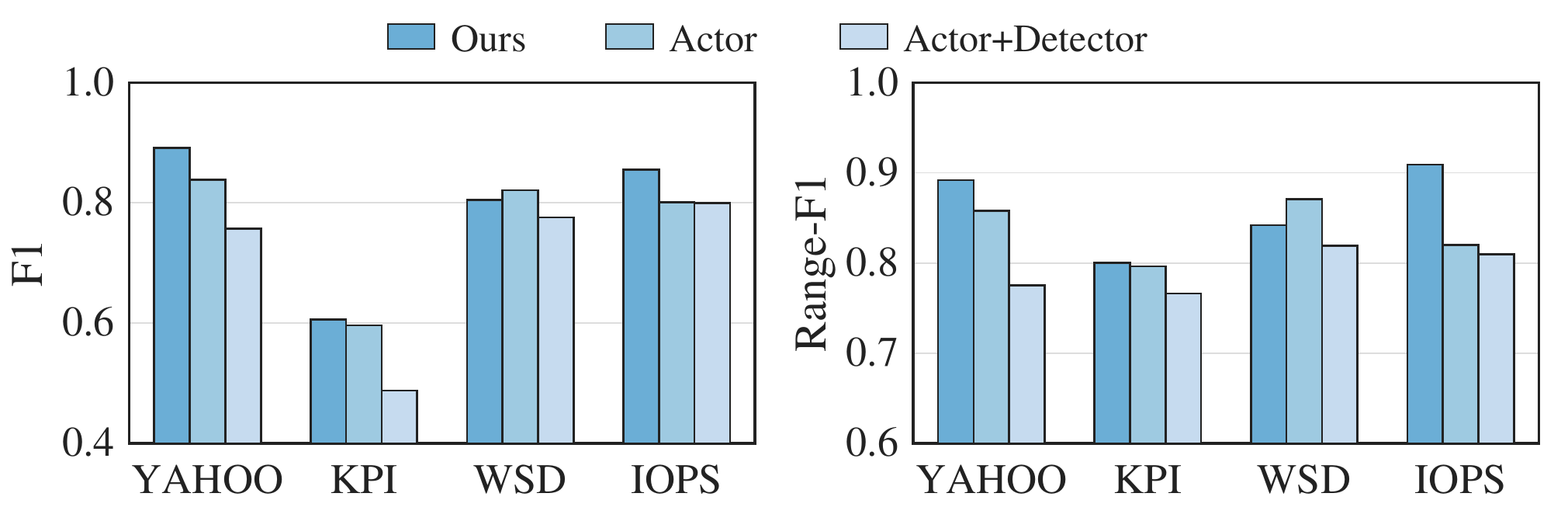}
\caption{Training configurations across roles.}
\Description{F1 and Range-F1 for Detector-only, Actor-only, and Actor plus Detector training across four datasets.}
\label{fig:training}
\end{figure}

\paragraph{Metrics.}
We report per-sequence averages of F1, best-F1, AUC-PR, and range-F1 to evaluate anomaly detection performance. Per-sequence averaging gives equal weight to each evaluated time series. F1 and range-F1 assess final binary predictions, while AUC-PR measures anomaly score ranking using ordinal confidence scores for AnomaMind. Best-F1 represents the optimal F1 obtained through oracle threshold selection using evaluation labels. Together, these metrics capture complementary aspects of anomaly localization quality. Further details on metric computation and aggregation are provided in Appendix~\ref{app:evaluation-metrics}.

\begin{table}[t]
  \centering
  \small
  \caption{Ablation study of the different diagnostic tool variants across diverse datasets.}
  \label{tab:ablation-tool}
  \footnotesize
  \setlength{\tabcolsep}{2.6pt}
  \resizebox{\linewidth}{!}{%
  \begin{tabular}{c| *{8}{c}}
  \toprule
  \multirow{2}{*}{Diagnostic Variants}
  & \multicolumn{2}{c}{YAHOO}
  & \multicolumn{2}{c}{KPI}
  & \multicolumn{2}{c}{WSD}
  & \multicolumn{2}{c}{IOPS} \\
  \cmidrule(lr){2-3} \cmidrule(lr){4-5} \cmidrule(lr){6-7} \cmidrule(lr){8-9}
  & F1 & Range-F1
  & F1 & Range-F1
  & F1 & Range-F1
  & F1 & Range-F1 \\
  \midrule
  w/o Feature Extraction
  & \underline{0.875} & \underline{0.871}
  & 0.569 & 0.738
  & 0.698 & 0.712
  & 0.842 & 0.820 \\
  w/o Value-based Detection
  & 0.863 & 0.863
  & 0.528 & 0.697
  & 0.647 & 0.693
  & 0.796 & 0.786 \\
  w/o Change-based Detection
  & 0.771 & 0.771
  & 0.587 & 0.764
  & \underline{0.735} & 0.724
  & 0.808 & 0.810 \\
  w/o Contextual Guidance
  & 0.845 & 0.845
  & \textbf{0.610} & \underline{0.784}
  & 0.720 & \underline{0.753}
  & \underline{0.845} & \underline{0.827} \\
  \midrule
  \rowcolor[gray]{0.95} \textbf{Ours}
  & \textbf{0.891} & \textbf{0.892}
  & \underline{0.606} & \textbf{0.800}
  & \textbf{0.805} & \textbf{0.842}
  & \textbf{0.855} & \textbf{0.909} \\
  \bottomrule
  \end{tabular}
  }
\end{table}

\begin{table}[t]
  \centering
  \small
  \caption{Impact of tool-use strategies on anomaly detection performance and efficiency.}
  \label{tab:tool-use-strategies}
  \footnotesize
  \setlength{\tabcolsep}{2.6pt}
  \resizebox{\linewidth}{!}{%
  \begin{tabular}{c| *{8}{c}}
  \toprule
  \multirow{2}{*}{\makecell{Tool-Use\\Strategies}}
  & \multicolumn{2}{c}{YAHOO}
  & \multicolumn{2}{c}{KPI}
  & \multicolumn{2}{c}{WSD}
  & \multicolumn{2}{c}{IOPS} \\
  \cmidrule(lr){2-3} \cmidrule(lr){4-5} \cmidrule(lr){6-7} \cmidrule(lr){8-9}
  & F1 $\uparrow$ & Calls $\downarrow$
  & F1 $\uparrow$ & Calls $\downarrow$
  & F1 $\uparrow$ & Calls $\downarrow$
  & F1 $\uparrow$ & Calls $\downarrow$ \\
  \midrule
  Fixed Tool
  & \underline{0.782} & \underline{3.44} & \underline{0.574} & \underline{3.95} & \underline{0.635} & \underline{3.57} & \underline{0.712} & \underline{3.44} \\
  Random Tool
  & 0.740 & \textbf{2.16} & 0.423 & \textbf{2.52} & 0.455 & \textbf{2.38} & 0.392 & \textbf{2.41} \\
  \midrule
  \rowcolor[gray]{0.95} \textbf{Ours}
  & \textbf{0.891} & 3.73 & \textbf{0.606} & 4.26 & \textbf{0.805} & 3.81 & \textbf{0.855} & 3.67 \\
  \bottomrule
  \end{tabular}
  }
\end{table}

\subsection{Overall Detection Performance}
\label{sec:overall-detection-performance}

As shown in Table~\ref{tab:main-new}, AnomaMind demonstrates competitive detection performance across all four benchmarks. The multi-domain model is jointly trained across all datasets, while separate in-domain models are trained for individual datasets. Both configurations outperform AnomaMind-Direct and show stronger overall performance than AnomaMind-Zero, although the latter remains competitive on certain metrics. These results support the benefits of combining diagnostic interactions with task-specific detector optimization, whose individual contributions are further examined in the ablation studies.

In-domain and multi-domain training exhibit different strengths across benchmarks. In-domain training performs better on some datasets, while multi-domain training excels on others, indicating that cross-domain learning provides dataset-dependent benefits. These patterns suggest that differences in temporal dynamics and anomaly characteristics influence the relative benefits of domain-specific versus shared representations. Certain baselines remain competitive in anomaly ranking and interval-level detection, highlighting the importance of evaluating detection quality beyond pointwise F1. Overall, multi-domain training achieves competitive performance across diverse datasets within a unified model.

\subsection{Diagnostic Workflow Analysis}
\label{sec:diagnostic-workflow-analysis}

\begin{table}[t]
  \centering
  \small
  \caption{Ablation study of SFT and RL contributions to anomaly detection performance.}
  \label{tab:ablation-training}
  \footnotesize
  \setlength{\tabcolsep}{2.6pt}
  \resizebox{\linewidth}{!}{%
  \begin{tabular}{c| *{8}{c}}
  \toprule
  \multirow{2}{*}{Training Strategies}
  & \multicolumn{2}{c}{YAHOO}
  & \multicolumn{2}{c}{KPI}
  & \multicolumn{2}{c}{WSD}
  & \multicolumn{2}{c}{IOPS} \\
  \cmidrule(lr){2-3} \cmidrule(lr){4-5} \cmidrule(lr){6-7} \cmidrule(lr){8-9}
  & F1 & Range-F1
  & F1 & Range-F1
  & F1 & Range-F1
  & F1 & Range-F1 \\
  \midrule
  w/o Both
  & 0.220 & 0.528
  & 0.065 & 0.424
  & 0.404 & 0.558
  & 0.400 & 0.673 \\
  w/o RL 
  & \underline{0.782} & \underline{0.807}
  & \underline{0.475} & \underline{0.724}
  & \underline{0.742} & \underline{0.777}
  & \underline{0.801} & \underline{0.820} \\
  w/o SFT
  & 0.607 & 0.661
  & 0.334 & 0.673
  & 0.691 & 0.690
  & 0.728 & 0.797 \\
  \midrule
  \rowcolor[gray]{0.95} \textbf{Ours}
  & \textbf{0.891} & \textbf{0.892}
  & \textbf{0.606} & \textbf{0.800}
  & \textbf{0.805} & \textbf{0.842}
  & \textbf{0.855} & \textbf{0.909} \\
  \bottomrule
  \end{tabular}
  }
\end{table}

\begin{figure}[t]
\centering
\includegraphics[width=\linewidth]{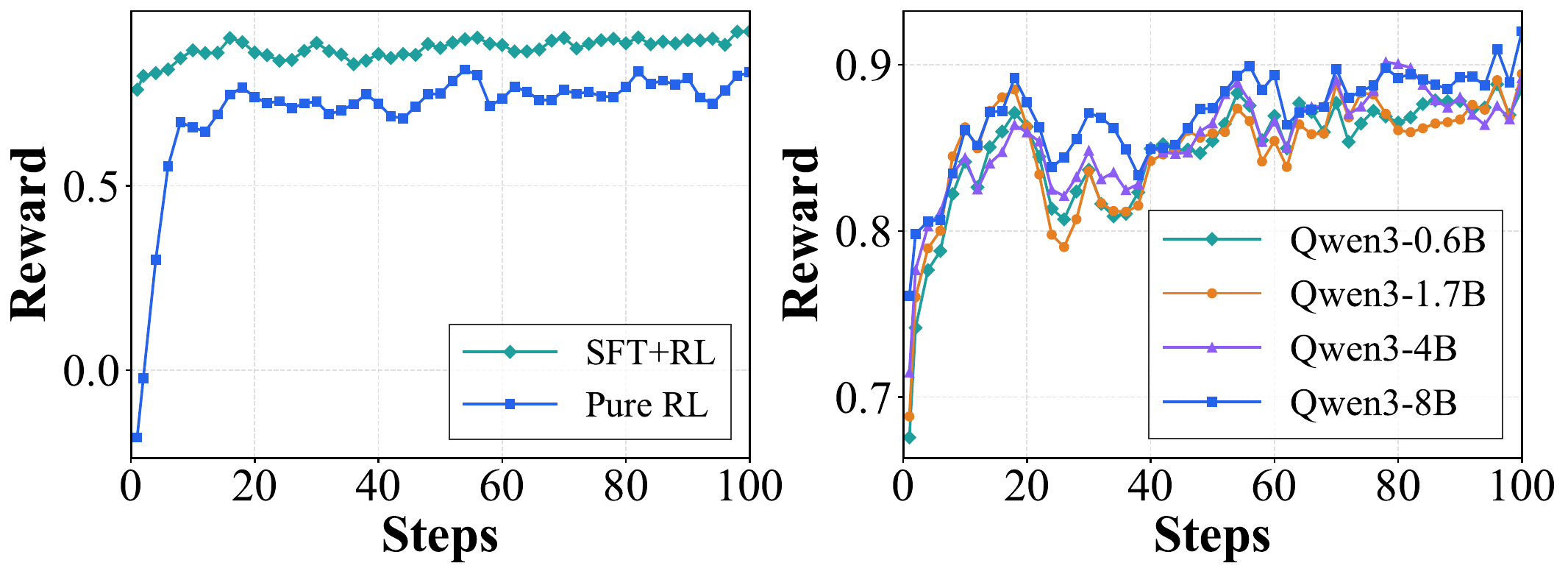}
\caption{Training reward trajectories for different initialization strategies and detector model sizes, shown in the left and right panels, respectively.}
\Description{Reward trajectories comparing supervised initialization followed by reinforcement learning with pure reinforcement learning, and comparing Detector sizes.}
\label{fig:step}
\end{figure}
\subsubsection{Workflow Roles}

As shown in Figure~\ref{fig:workflow}, removing the locator consistently reduces performance across all datasets, highlighting the importance of initial localization in guiding subsequent diagnosis by narrowing the search space before more detailed numerical inspection. Removing the evaluator also decreases F1. These results demonstrate the effectiveness of the proposed diagnostic workflow, where initial localization guides targeted numerical evidence acquisition and self-reflection refines anomaly decisions. The performance degradation caused by removing either role further highlights the importance of coordinating different stages to support accurate anomaly localization.

\subsubsection{Training Configurations Across Workflow Roles}

Figure~\ref{fig:training} compares different training configurations across workflow roles and examines their effects on anomaly detection performance. The proposed hybrid inference configuration achieves the strongest detection performance on most benchmarks, while training the actor alone or jointly optimizing both roles does not yield consistent improvements. These results support the effectiveness of retaining general-purpose reasoning capabilities for adaptive numerical evidence acquisition while optimizing a specialized LLM detector for evidence-grounded interval prediction. As the configurations involve different model assignments, the comparison reflects system-level effectiveness rather than isolating the effects of individual training roles.

\begin{table}[t]
  \centering
  \small
  \caption{Experimental results with different general-purpose model variants under the proposed framework.}
  \label{tab:model-comparison}
  \resizebox{\linewidth}{!}{%
  \begin{tabular}{c|cccccccc}
  \toprule
  \multirow{2}{*}{\makecell{General-Purpose\\LLMs}}
  & \multicolumn{2}{c}{YAHOO}
  & \multicolumn{2}{c}{KPI}
  & \multicolumn{2}{c}{WSD}
  & \multicolumn{2}{c}{IOPS} \\
  \cmidrule(lr){2-3} \cmidrule(lr){4-5} \cmidrule(lr){6-7} \cmidrule(lr){8-9}
  & F1 & Range-F1
  & F1 & Range-F1
  & F1 & Range-F1
  & F1 & Range-F1 \\
  \midrule
  Gemini-3.1-flash
  & \textbf{0.891} & \textbf{0.892}
  & \textbf{0.606} & \textbf{0.800}
  & \underline{0.805} & \underline{0.842}
  & \underline{0.855} & \textbf{0.909} \\
  grok-4-1-fast
  & \underline{0.848} & \underline{0.860}
  & 0.424 & \underline{0.749}
  & \textbf{0.836} & \textbf{0.856}
  & \textbf{0.871} & \textbf{0.909} \\
  GPT-5-mini
  & 0.774 & 0.853
  & \underline{0.447} & 0.691
  & 0.766 & 0.840
  & 0.740 & \underline{0.796} \\
  \bottomrule
  \end{tabular}
  }
\end{table}

\begin{table}[t]
  \centering
  \caption{Impact of specialized detector size on anomaly detection performance.}
  \label{tab:model-size}
  \footnotesize
  \resizebox{\linewidth}{!}{%
  \begin{tabular}{c|cccccccc}
  \toprule
  \multirow{2}{*}{Detector Sizes}
  & \multicolumn{2}{c}{YAHOO}
  & \multicolumn{2}{c}{KPI}
  & \multicolumn{2}{c}{WSD}
  & \multicolumn{2}{c}{IOPS} \\
  \cmidrule(lr){2-3} \cmidrule(lr){4-5} \cmidrule(lr){6-7} \cmidrule(lr){8-9}
  & F1 & Range-F1
  & F1 & Range-F1
  & F1 & Range-F1
  & F1 & Range-F1 \\
  \midrule
  Qwen3-0.6B
  & 0.827 & 0.833
  & 0.533 & 0.737
  & 0.705 & 0.763
  & 0.806 & 0.811 \\
  Qwen3-1.7B
  & 0.852 & 0.854
  & 0.523 & 0.751
  & 0.699 & 0.758
  & 0.803 & 0.807 \\
  Qwen3-4B
  & \underline{0.870} & \underline{0.867}
  & \underline{0.575} & \textbf{0.802}
  & \underline{0.728} & \underline{0.767}
  & \underline{0.837} & \underline{0.895} \\
  \midrule
  \rowcolor[gray]{0.95} Qwen3-8B
  & \textbf{0.891} & \textbf{0.892}
  & \textbf{0.606} & \underline{0.800}
  & \textbf{0.805} & \textbf{0.842}
  & \textbf{0.855} & \textbf{0.909} \\
  \bottomrule
  \end{tabular}
  }
\end{table}

\subsubsection{Diagnostic Tools}

As shown in Table~\ref{tab:ablation-tool}, the complete tool configuration achieves the highest range-F1 across all benchmarks. Removing individual diagnostic components generally reduces detection performance, with value-based and change-based tools showing particularly pronounced effects on certain datasets. Statistical feature extraction and contextual guidance also contribute to interval-level detection, although their effects on F1 vary across datasets. These results highlight the complementary contributions of different diagnostic evidence sources within the proposed workflow. While these ablations examine the availability of diagnostic evidence,
the following experiments investigate the effectiveness of different tool-use strategies.

\subsubsection{Tool-Use Strategies}
\label{sec:tool-use-strategies}
Table~\ref{tab:tool-use-strategies} compares different tool-use strategies under consistent experimental settings. Fixed follows predefined diagnostic rules, while random introduces stochastic tool selection and threshold adjustments. In contrast, adaptive dynamically selects diagnostic tools, adjusts thresholds, and refines query regions based on accumulated evidence, allowing the diagnostic process to focus on suspicious patterns and progressively refine anomaly boundaries. Adaptive consistently achieves superior detection performance with only modest additional computational overhead. These results demonstrate that evidence-guided tool selection and iterative diagnosis are more effective than predefined or random strategies, highlighting the importance of adaptive evidence acquisition for accurate anomaly localization.

\begin{figure*}[t]
\centering
\includegraphics[width=0.95\linewidth]{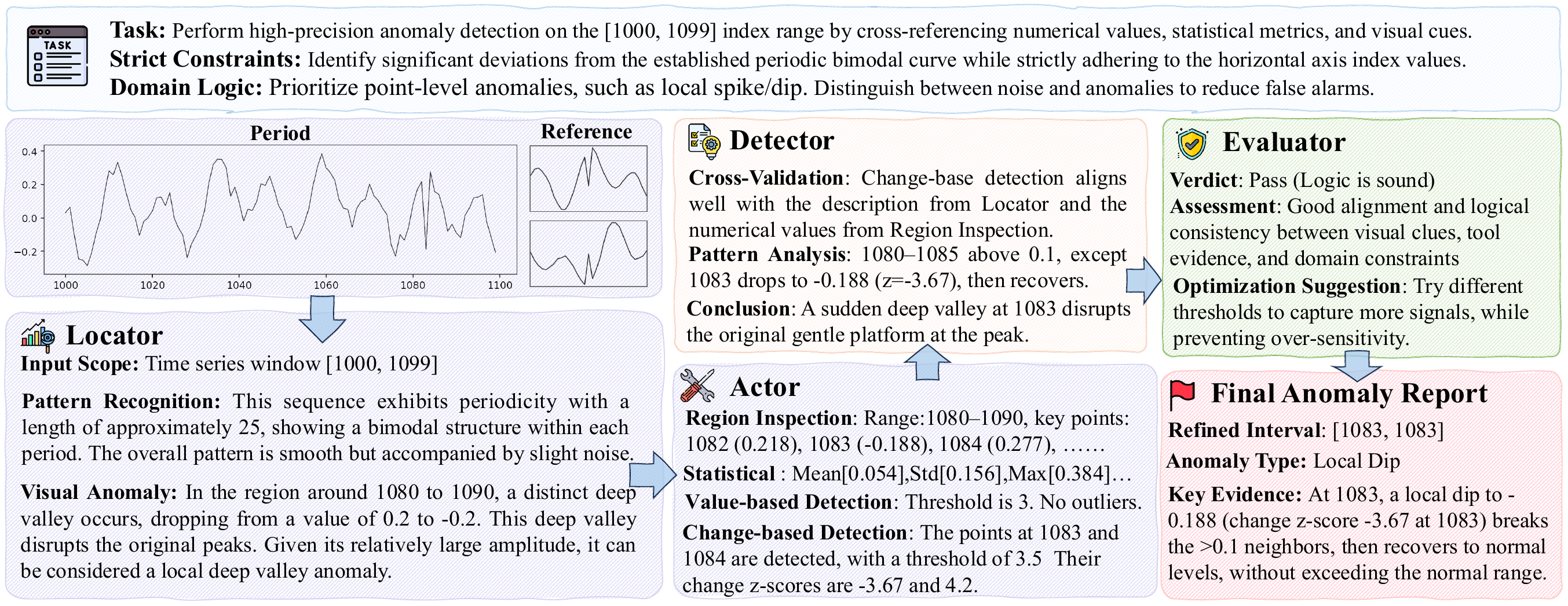}
\caption{An illustrative example of a coarse-to-fine diagnostic workflow for agentic time series anomaly detection.}
\Description{A Locator proposes a suspicious valley, an Actor gathers numerical evidence, a Detector localizes a brief dip, and an Evaluator checks consistency.}
\label{fig:case}
\end{figure*}

\subsection{Hybrid Inference and Optimization}
\label{sec:hybrid-inference-analysis}

\subsubsection{Training Strategy}

As shown in Table~\ref{tab:ablation-training}, the full training configuration achieves the highest average F1 and range-F1 across the evaluated datasets. SFT alone retains much of the full model's performance, whereas RL without supervised initialization yields substantially lower scores, highlighting the importance of SFT for learning to interpret diagnostic evidence. Incorporating RL further improves detection performance, although the gains vary across datasets and metrics. The variant without either training stage performs worst, supporting the effectiveness of task-specific optimization in learning the mapping from diagnostic evidence to anomaly decisions.
Figure~\ref{fig:step} further illustrates the corresponding training reward trajectories.

\subsubsection{Training Dynamics}
\label{sec:training-trajectories}
Figure~\ref{fig:step} presents the training reward trajectories under two experimental settings. In the left panel, supervised initialization maintains higher rewards throughout training, whereas RL without SFT improves rapidly in the early stages but remains at a lower reward level. In the right panel, detectors of different sizes exhibit similar upward trends, with larger models generally achieving higher rewards despite fluctuations and overlapping trajectories. These observations highlight the benefits of supervised initialization for RL optimization and illustrate how detector capacity influences reward dynamics during training.

\subsection{Model Configuration and Efficiency}
\label{sec:model-inference-analysis}

\subsubsection{Detector Model Size}
\label{sec:model-size-analysis}

Table~\ref{tab:model-size} examines the impact of detector size on anomaly detection performance. Qwen3-8B achieves the strongest overall performance, although smaller models remain competitive on certain metrics, indicating that increased model capacity does not consistently translate into better detection results. Together with Section~\ref{sec:inference-efficiency}, these results show the trade-off between computational cost and detection quality.

\subsubsection{General-Purpose Model Variants}
\label{sec:general-model-analysis}

Table~\ref{tab:model-comparison} examines the impact of frozen general-purpose model selection on anomaly detection performance. Gemini-3.1-flash performs competitively across benchmarks, although its relative performance varies by dataset. GPT-5-mini generally achieves lower performance, although it remains competitive in certain settings. We retain Gemini-3.1-flash as the default backbone for consistency. These results highlight the influence of general-purpose model selection on detection quality.

\subsubsection{Sampling-Temperature Sensitivity}
\label{sec:temperature-analysis}
\begin{figure}[!t]
\centering
\includegraphics[width=\linewidth]{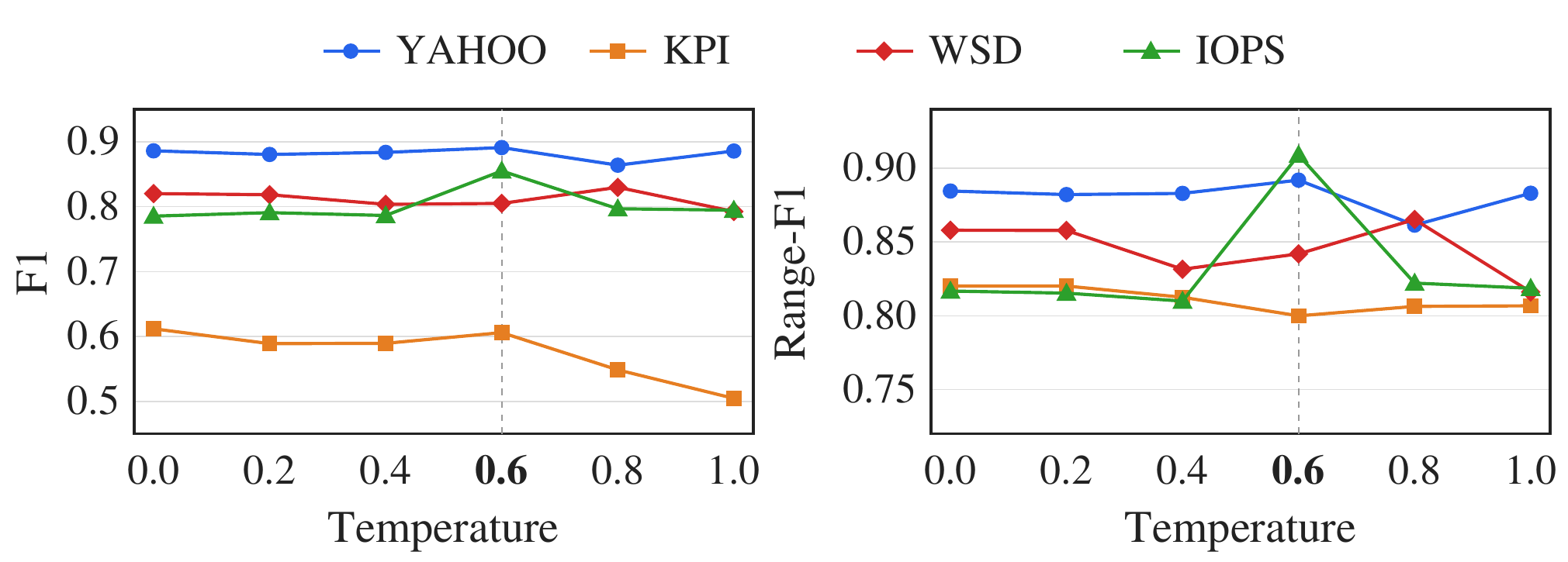}
\caption{Detector-temperature sensitivity. Dashed lines mark the default temperature $T=0.6$.}
\label{fig:temperature sensitivity}
\end{figure}
\begin{table}[t]
\centering
\small
\caption{Inference efficiency by detector size and concurrency.}
\label{tab:detector-efficiency}
\setlength{\tabcolsep}{2pt}
\renewcommand{\arraystretch}{1.0}
\begin{tabular*}{\linewidth}{@{\extracolsep{\fill}}lrrrrr@{}}
\toprule
\makecell[l]{Detectors\\Memory}
& Conc.
& \makecell[r]{Thpt.\\requests/s}
& \makecell[r]{Mean lat.\\ms}
& \makecell[r]{TTFT\\ms}
& \makecell[r]{Decode\\tokens/s} \\
\midrule
\multirow{3}{*}{\makecell[l]{Qwen3-0.6B\\1.12\,GB}}
& 32  & 6.26  & 4,939 & 157 & 977 \\
& 64  & 9.90  & 6,137 & 245 & 1,543 \\
& 128 & 12.46 & 9,593 & 598 & 1,943 \\
\midrule
\multirow{3}{*}{\makecell[l]{Qwen3-8B\\15.27\,GB}}
& 32  & 3.53 & 8,854  & 320   & 550 \\
& 64  & 4.45 & 13,965 & 646   & 684 \\
& 128 & 4.95 & 24,951 & 1,626 & 771 \\
\bottomrule
\end{tabular*}
\end{table}

Figure~\ref{fig:temperature sensitivity} examines the sensitivity of anomaly detection performance to detector sampling temperature. The results exhibit varying trends across benchmarks, with some showing relatively stable performance and others displaying greater fluctuations. Both F1 and range-F1 are affected by temperature selection, with no single setting consistently achieving the best performance.
The dashed line indicates the default temperature of $T=0.6$.

\subsubsection{Inference Efficiency Analysis}
\label{sec:inference-efficiency}
Table~\ref{tab:detector-efficiency} compares inference efficiency across detector sizes and concurrency levels (Conc.). Throughput (Thpt.) and mean latency (Mean lat.) measure workflow-level efficiency, while memory usage, time to first token (TTFT), and decoding speed characterize detector serving. Smaller detectors achieve lower latency and higher throughput. Increasing concurrency improves throughput at the cost of latency, with diminishing gains for larger detectors, highlighting the trade-off between detection quality and efficiency.

\subsection{Qualitative Analysis}
\label{sec:qualitative-analysis}
Figure~\ref{fig:case} illustrates evidence-grounded anomaly localization on a periodic bimodal window. The locator identifies a suspicious valley, where value-based inspection finds no clear amplitude outlier, while change-based analysis reveals sharp transitions. Regional inspection further confirms the subsequent recovery. The detector narrows the broader candidate region to $[1083,1083]$, and the evaluator accepts the result after reviewing the consistency of the reported evidence. This case illustrates coarse-to-fine interval refinement followed by the evaluator's consistency review.

\Needspace{10\baselineskip}
\section{Conclusion}

In this work, we presented AnomaMind, a coarse-to-fine agentic diagnosis framework for precise anomaly interval localization in web service KPI analysis. AnomaMind integrates initial localization, adaptive evidence acquisition through multi-turn tool interactions, and self-reflective refinement, supported by a tool engine combining knowledge memory and numerical analysis. A hybrid inference strategy employs frozen general-purpose LLMs to coordinate the diagnostic workflow, while a specialized detector is optimized through SFT and RL for evidence-grounded interval prediction. Experiments on four benchmarks spanning web traffic, service KPIs, and system-monitoring signals demonstrate competitive performance under both in-domain and multi-domain training settings. Further analyses validate the effectiveness of targeted evidence acquisition and iterative refinement, while highlighting the trade-off between detection quality and computational efficiency. Future work will explore adaptive evidence acquisition under limited budgets and generalization across diverse KPI patterns.

\bibliographystyle{ACM-Reference-Format}
\bibliography{main}


\begin{thebibliography}{51}


\ifx \showCODEN    \undefined \def \showCODEN     #1{\unskip}     \fi
\ifx \showISBNx    \undefined \def \showISBNx     #1{\unskip}     \fi
\ifx \showISBNxiii \undefined \def \showISBNxiii  #1{\unskip}     \fi
\ifx \showISSN     \undefined \def \showISSN      #1{\unskip}     \fi
\ifx \showLCCN     \undefined \def \showLCCN      #1{\unskip}     \fi
\ifx \shownote     \undefined \def \shownote      #1{#1}          \fi
\ifx \showarticletitle \undefined \def \showarticletitle #1{#1}   \fi
\ifx \showURL      \undefined \def \showURL       {\relax}        \fi
\providecommand\bibfield[2]{#2}
\providecommand\bibinfo[2]{#2}
\providecommand\natexlab[1]{#1}
\providecommand\showeprint[2][]{arXiv:#2}

\bibitem[Alnegheimish et~al\mbox{.}(2024)]%
        {alnegheimish2024large}
\bibfield{author}{\bibinfo{person}{Sarah Alnegheimish}, \bibinfo{person}{Linh
  Nguyen}, \bibinfo{person}{Laure Berti-Equille}, {and} \bibinfo{person}{Kalyan
  Veeramachaneni}.} \bibinfo{year}{2024}\natexlab{}.
\newblock \showarticletitle{Large language models can be zero-shot anomaly
  detectors for time series?}
\newblock \bibinfo{journal}{\emph{arXiv preprint arXiv:2405.14755}}
  (\bibinfo{year}{2024}).
\newblock


\bibitem[Ansari et~al\mbox{.}(2024)]%
        {ansari2024chronos}
\bibfield{author}{\bibinfo{person}{Abdul~Fatir Ansari},
  \bibinfo{person}{Lorenzo Stella}, \bibinfo{person}{Caner Turkmen},
  \bibinfo{person}{Xiyuan Zhang}, \bibinfo{person}{Pedro Mercado},
  \bibinfo{person}{Huibin Shen}, \bibinfo{person}{Oleksandr Shchur},
  \bibinfo{person}{Syama~Sundar Rangapuram}, \bibinfo{person}{Sebastian~Pineda
  Arango}, \bibinfo{person}{Shubham Kapoor}, {et~al\mbox{.}}}
  \bibinfo{year}{2024}\natexlab{}.
\newblock \showarticletitle{Chronos: Learning the language of time series}.
\newblock \bibinfo{journal}{\emph{arXiv preprint arXiv:2403.07815}}
  (\bibinfo{year}{2024}).
\newblock


\bibitem[Audibert et~al\mbox{.}(2020)]%
        {audibert2020usad}
\bibfield{author}{\bibinfo{person}{Julien Audibert}, \bibinfo{person}{Pietro
  Michiardi}, \bibinfo{person}{Fr{\'e}d{\'e}ric Guyard},
  \bibinfo{person}{S{\'e}bastien Marti}, {and} \bibinfo{person}{Maria~A
  Zuluaga}.} \bibinfo{year}{2020}\natexlab{}.
\newblock \showarticletitle{Usad: Unsupervised anomaly detection on
  multivariate time series}. In \bibinfo{booktitle}{\emph{Proceedings of the
  26th ACM SIGKDD International Conference on Knowledge Discovery \& Data
  Mining}}. \bibinfo{pages}{3395--3404}.
\newblock


\bibitem[Breunig et~al\mbox{.}(2000)]%
        {breunig2000lof}
\bibfield{author}{\bibinfo{person}{Markus~M Breunig},
  \bibinfo{person}{Hans-Peter Kriegel}, \bibinfo{person}{Raymond~T Ng}, {and}
  \bibinfo{person}{J{\"o}rg Sander}.} \bibinfo{year}{2000}\natexlab{}.
\newblock \showarticletitle{LOF: identifying density-based local outliers}. In
  \bibinfo{booktitle}{\emph{Proceedings of the 2000 ACM SIGMOD International
  Conference on Management of Data}}. \bibinfo{pages}{93--104}.
\newblock


\bibitem[Carmona et~al\mbox{.}(2022)]%
        {carmona2022neural}
\bibfield{author}{\bibinfo{person}{Chris~U. Carmona},
  \bibinfo{person}{Fran{\c{c}}ois-Xavier Aubet}, \bibinfo{person}{Valentin
  Flunkert}, {and} \bibinfo{person}{Jan Gasthaus}.}
  \bibinfo{year}{2022}\natexlab{}.
\newblock \showarticletitle{Neural Contextual Anomaly Detection for Time
  Series}. In \bibinfo{booktitle}{\emph{Proceedings of the Thirty-First
  International Joint Conference on Artificial Intelligence}}.
  \bibinfo{pages}{2843--2851}.
\newblock


\bibitem[Chandola et~al\mbox{.}(2009)]%
        {chandola2009anomaly}
\bibfield{author}{\bibinfo{person}{Varun Chandola}, \bibinfo{person}{Arindam
  Banerjee}, {and} \bibinfo{person}{Vipin Kumar}.}
  \bibinfo{year}{2009}\natexlab{}.
\newblock \showarticletitle{Anomaly detection: A survey}.
\newblock \bibinfo{journal}{\emph{Comput. Surveys}} \bibinfo{volume}{41},
  \bibinfo{number}{3} (\bibinfo{year}{2009}), \bibinfo{pages}{1--58}.
\newblock


\bibitem[Chang et~al\mbox{.}(2026)]%
        {chang2026survey}
\bibfield{author}{\bibinfo{person}{Ching Chang}, \bibinfo{person}{Yidan Shi},
  \bibinfo{person}{Defu Cao}, \bibinfo{person}{Wei Yang},
  \bibinfo{person}{Jeehyun Hwang}, \bibinfo{person}{Haixin Wang},
  \bibinfo{person}{Jiacheng Pang}, \bibinfo{person}{Wei Wang},
  \bibinfo{person}{Yan Liu}, \bibinfo{person}{Wen-Chih Peng}, {and}
  \bibinfo{person}{Tien-Fu Chen}.} \bibinfo{year}{2026}\natexlab{}.
\newblock \showarticletitle{A Survey of Reasoning and Agentic Systems in Time
  Series with Large Language Models}.
\newblock \bibinfo{journal}{\emph{Transactions on Machine Learning Research}}
  (\bibinfo{year}{2026}).
\newblock
\shownote{Survey Certification}.
\newblock
\showISSN{2835-8856}


\bibitem[Das et~al\mbox{.}(2024)]%
        {das2024decoder}
\bibfield{author}{\bibinfo{person}{Abhimanyu Das}, \bibinfo{person}{Weihao
  Kong}, \bibinfo{person}{Rajat Sen}, {and} \bibinfo{person}{Yichen Zhou}.}
  \bibinfo{year}{2024}\natexlab{}.
\newblock \showarticletitle{A decoder-only foundation model for time-series
  forecasting}. In \bibinfo{booktitle}{\emph{Proceedings of the 41st
  International Conference on Machine Learning}}.
\newblock


\bibitem[Deforce et~al\mbox{.}(2025)]%
        {deforce2025tsap}
\bibfield{author}{\bibinfo{person}{Boje Deforce}, \bibinfo{person}{Meng-Chieh
  Lee}, \bibinfo{person}{Bart Baesens}, \bibinfo{person}{Estefan{\'i}a
  Serral~Asensio}, \bibinfo{person}{Jaemin Yoo}, {and} \bibinfo{person}{Leman
  Akoglu}.} \bibinfo{year}{2025}\natexlab{}.
\newblock \showarticletitle{End-To-End Self-Tuning Self-Supervised Time Series
  Anomaly Detection}. In \bibinfo{booktitle}{\emph{Proceedings of the 2025 SIAM
  International Conference on Data Mining}}. \bibinfo{publisher}{SIAM},
  \bibinfo{pages}{568--577}.
\newblock


\bibitem[Deng and Hooi(2021)]%
        {deng2021graph}
\bibfield{author}{\bibinfo{person}{Ailin Deng} {and} \bibinfo{person}{Bryan
  Hooi}.} \bibinfo{year}{2021}\natexlab{}.
\newblock \showarticletitle{Graph Neural Network-Based Anomaly Detection in
  Multivariate Time Series}. In \bibinfo{booktitle}{\emph{Proceedings of the
  AAAI Conference on Artificial Intelligence}}, Vol.~\bibinfo{volume}{35}.
  \bibinfo{pages}{4027--4035}.
\newblock


\bibitem[Gu et~al\mbox{.}(2025)]%
        {gu2025argos}
\bibfield{author}{\bibinfo{person}{Yile Gu}, \bibinfo{person}{Yifan Xiong},
  \bibinfo{person}{Jonathan Mace}, \bibinfo{person}{Yuting Jiang},
  \bibinfo{person}{Yigong Hu}, \bibinfo{person}{Baris Kasikci}, {and}
  \bibinfo{person}{Peng Cheng}.} \bibinfo{year}{2025}\natexlab{}.
\newblock \showarticletitle{Argos: Agentic time-series anomaly detection with
  autonomous rule generation via large language models}.
\newblock \bibinfo{journal}{\emph{arXiv preprint arXiv:2501.14170}}
  (\bibinfo{year}{2025}).
\newblock


\bibitem[Guan et~al\mbox{.}(2026)]%
        {guan2026timeomni}
\bibfield{author}{\bibinfo{person}{Tong Guan}, \bibinfo{person}{Zijie Meng},
  \bibinfo{person}{Dianqi Li}, \bibinfo{person}{Shiyu Wang},
  \bibinfo{person}{Chao-Han~Huck Yang}, \bibinfo{person}{Qingsong Wen},
  \bibinfo{person}{Zuozhu Liu}, \bibinfo{person}{Sabato Siniscalchi},
  \bibinfo{person}{Ming Jin}, {and} \bibinfo{person}{Shirui Pan}.}
  \bibinfo{year}{2026}\natexlab{}.
\newblock \showarticletitle{{TimeOmni-1}: Incentivizing Complex Reasoning with
  Time Series in Large Language Models}. In
  \bibinfo{booktitle}{\emph{International Conference on Learning
  Representations}}.
\newblock


\bibitem[He et~al\mbox{.}(2026)]%
        {he2026harnessing}
\bibfield{author}{\bibinfo{person}{Zelin He}, \bibinfo{person}{Sarah
  Alnegheimish}, {and} \bibinfo{person}{Matthew Reimherr}.}
  \bibinfo{year}{2026}\natexlab{}.
\newblock \showarticletitle{Harnessing vision-language models for time series
  anomaly detection}. In \bibinfo{booktitle}{\emph{Proceedings of the AAAI
  Conference on Artificial Intelligence}}, Vol.~\bibinfo{volume}{40}.
  \bibinfo{pages}{21690--21698}.
\newblock


\bibitem[Huang et~al\mbox{.}(2026)]%
        {huang2026hydra}
\bibfield{author}{\bibinfo{person}{Mingyi Huang}, \bibinfo{person}{Qinghua
  Liu}, \bibinfo{person}{Paul Boniol}, {and} \bibinfo{person}{John
  Paparrizos}.} \bibinfo{year}{2026}\natexlab{}.
\newblock \showarticletitle{{HYDRA}: A Multi-Level Hierarchy-Driven Approach
  for Robust Anomaly Detection in Time Series}.
\newblock \bibinfo{journal}{\emph{Proceedings of the ACM on Management of
  Data}} \bibinfo{volume}{4}, \bibinfo{number}{3}, Article
  \bibinfo{articleno}{197} (\bibinfo{date}{June} \bibinfo{year}{2026}),
  \bibinfo{numpages}{30}~pages.
\newblock


\bibitem[Kang et~al\mbox{.}(2026)]%
        {kang2026sage}
\bibfield{author}{\bibinfo{person}{Hyeongwon Kang}, \bibinfo{person}{Jeongseob
  Kim}, \bibinfo{person}{Jinwoo Park}, {and} \bibinfo{person}{Pilsung Kang}.}
  \bibinfo{year}{2026}\natexlab{}.
\newblock \showarticletitle{Detecting Time Series Anomalies Like an Expert: A
  Multi-Agent {LLM} Framework with Specialized Analyzers}.
\newblock \bibinfo{journal}{\emph{arXiv preprint arXiv:2605.05725}}
  (\bibinfo{year}{2026}).
\newblock


\bibitem[Kim et~al\mbox{.}(2024)]%
        {kim2024model}
\bibfield{author}{\bibinfo{person}{Dongmin Kim}, \bibinfo{person}{Sunghyun
  Park}, {and} \bibinfo{person}{Jaegul Choo}.} \bibinfo{year}{2024}\natexlab{}.
\newblock \showarticletitle{When model meets new normals: Test-time adaptation
  for unsupervised time-series anomaly detection}. In
  \bibinfo{booktitle}{\emph{Proceedings of the AAAI Conference on Artificial
  Intelligence}}, Vol.~\bibinfo{volume}{38}. \bibinfo{pages}{13113--13121}.
\newblock


\bibitem[Laptev et~al\mbox{.}(2015)]%
        {laptev2015generic}
\bibfield{author}{\bibinfo{person}{Nikolay Laptev}, \bibinfo{person}{Saeed
  Amizadeh}, {and} \bibinfo{person}{Ian Flint}.}
  \bibinfo{year}{2015}\natexlab{}.
\newblock \showarticletitle{Generic and scalable framework for automated
  time-series anomaly detection}. In \bibinfo{booktitle}{\emph{Proceedings of
  the 21st ACM SIGKDD International Conference on Knowledge Discovery and
  \mbox{Data Mining}}}. \bibinfo{pages}{1939--1947}.
\newblock


\bibitem[Liu et~al\mbox{.}(2024)]%
        {liu2024large}
\bibfield{author}{\bibinfo{person}{Chen Liu}, \bibinfo{person}{Shibo He},
  \bibinfo{person}{Qihang Zhou}, \bibinfo{person}{Shizhong Li}, {and}
  \bibinfo{person}{Wenchao Meng}.} \bibinfo{year}{2024}\natexlab{}.
\newblock \showarticletitle{Large language model guided knowledge distillation
  for time series anomaly detection}. In \bibinfo{booktitle}{\emph{Proceedings
  of the Thirty-Third International Joint Conference on \mbox{Artificial
  Intelligence}}}. \bibinfo{pages}{2162--2170}.
\newblock


\bibitem[Liu et~al\mbox{.}(2015)]%
        {liu2015opprentice}
\bibfield{author}{\bibinfo{person}{Dapeng Liu}, \bibinfo{person}{Youjian Zhao},
  \bibinfo{person}{Haowen Xu}, \bibinfo{person}{Yongqian Sun},
  \bibinfo{person}{Dan Pei}, \bibinfo{person}{Jiao Luo},
  \bibinfo{person}{Xiaowei Jing}, {and} \bibinfo{person}{Mei Feng}.}
  \bibinfo{year}{2015}\natexlab{}.
\newblock \showarticletitle{{Opprentice}: Towards Practical and Automatic
  Anomaly Detection Through Machine Learning}. In
  \bibinfo{booktitle}{\emph{Proceedings of the 2015 Internet Measurement
  Conference}}. \bibinfo{publisher}{ACM}, \bibinfo{pages}{211--224}.
\newblock


\bibitem[Liu et~al\mbox{.}(2025)]%
        {liu2025large}
\bibfield{author}{\bibinfo{person}{Jun Liu}, \bibinfo{person}{Chaoyun Zhang},
  \bibinfo{person}{Jiaxu Qian}, \bibinfo{person}{Minghua Ma},
  \bibinfo{person}{Si Qin}, \bibinfo{person}{Chetan Bansal},
  \bibinfo{person}{Qingwei Lin}, \bibinfo{person}{Saravan Rajmohan}, {and}
  \bibinfo{person}{Dongmei Zhang}.} \bibinfo{year}{2025}\natexlab{}.
\newblock \showarticletitle{Large language models can deliver accurate and
  interpretable time series anomaly detection}. In
  \bibinfo{booktitle}{\emph{Proceedings of the 31st ACM SIGKDD Conference on
  Knowledge Discovery and Data Mining V. 2}}. \bibinfo{pages}{4623--4634}.
\newblock


\bibitem[Liu and Paparrizos(2024)]%
        {liu2024elephant}
\bibfield{author}{\bibinfo{person}{Qinghua Liu} {and} \bibinfo{person}{John
  Paparrizos}.} \bibinfo{year}{2024}\natexlab{}.
\newblock \showarticletitle{The Elephant in the Room: Towards A Reliable
  Time-Series Anomaly Detection Benchmark}. In
  \bibinfo{booktitle}{\emph{Advances in Neural Information Processing
  Systems}}, Vol.~\bibinfo{volume}{37}.
\newblock


\bibitem[Malhotra et~al\mbox{.}(2015)]%
        {malhotra2015long}
\bibfield{author}{\bibinfo{person}{Pankaj Malhotra}, \bibinfo{person}{Lovekesh
  Vig}, \bibinfo{person}{Gautam Shroff}, \bibinfo{person}{Puneet Agarwal},
  {et~al\mbox{.}}} \bibinfo{year}{2015}\natexlab{}.
\newblock \showarticletitle{Long short term memory networks for anomaly
  detection in time series}. In \bibinfo{booktitle}{\emph{Proceedings of the
  23rd European Symposium on Artificial Neural Networks, Computational
  Intelligence and Machine Learning}}. \bibinfo{pages}{89--94}.
\newblock


\bibitem[Munir et~al\mbox{.}(2018)]%
        {munir2018deepant}
\bibfield{author}{\bibinfo{person}{Mohsin Munir}, \bibinfo{person}{Shoaib~Ahmed
  Siddiqui}, \bibinfo{person}{Andreas Dengel}, {and} \bibinfo{person}{Sheraz
  Ahmed}.} \bibinfo{year}{2018}\natexlab{}.
\newblock \showarticletitle{DeepAnT: A deep learning approach for unsupervised
  anomaly detection in time series}.
\newblock \bibinfo{journal}{\emph{IEEE Access}}  \bibinfo{volume}{7}
  (\bibinfo{year}{2018}), \bibinfo{pages}{1991--2005}.
\newblock


\bibitem[Nam et~al\mbox{.}(2024)]%
        {nam2024dualtf}
\bibfield{author}{\bibinfo{person}{Youngeun Nam}, \bibinfo{person}{Susik Yoon},
  \bibinfo{person}{Yooju Shin}, \bibinfo{person}{Minyoung Bae},
  \bibinfo{person}{Hwanjun Song}, \bibinfo{person}{Jae-Gil Lee}, {and}
  \bibinfo{person}{Byung~Suk Lee}.} \bibinfo{year}{2024}\natexlab{}.
\newblock \showarticletitle{Breaking the Time-Frequency Granularity Discrepancy
  in Time-Series Anomaly Detection}. In \bibinfo{booktitle}{\emph{Proceedings
  of the ACM on Web Conference 2024}}. \bibinfo{publisher}{ACM},
  \bibinfo{address}{New York, NY, USA}, \bibinfo{pages}{4204--4215}.
\newblock


\bibitem[Paparrizos et~al\mbox{.}(2022)]%
        {paparrizos2022volume}
\bibfield{author}{\bibinfo{person}{John Paparrizos}, \bibinfo{person}{Paul
  Boniol}, \bibinfo{person}{Themis Palpanas}, \bibinfo{person}{Ruey~S Tsay},
  \bibinfo{person}{Aaron~J Elmore}, {and} \bibinfo{person}{Michael~J
  Franklin}.} \bibinfo{year}{2022}\natexlab{}.
\newblock \showarticletitle{Volume Under the Surface: A New Accuracy Evaluation
  Measure for Time-Series Anomaly Detection.}
\newblock \bibinfo{journal}{\emph{Proceedings of the VLDB Endowment}}
  \bibinfo{volume}{15}, \bibinfo{number}{11} (\bibinfo{year}{2022}),
  \bibinfo{pages}{2774--2787}.
\newblock


\bibitem[Park et~al\mbox{.}(2025)]%
        {parkdelving}
\bibfield{author}{\bibinfo{person}{Junwoo Park}, \bibinfo{person}{Kyudan Jung},
  \bibinfo{person}{Dohyun Lee}, \bibinfo{person}{Hyuck Lee},
  \bibinfo{person}{Daehoon Gwak}, \bibinfo{person}{ChaeHun Park},
  \bibinfo{person}{Jaegul Choo}, {and} \bibinfo{person}{Jaewoong Cho}.}
  \bibinfo{year}{2025}\natexlab{}.
\newblock \showarticletitle{Delving into Large Language Models for Effective
  Time-Series Anomaly Detection}. In \bibinfo{booktitle}{\emph{Advances in
  Neural Information Processing Systems}}, Vol.~\bibinfo{volume}{38}.
\newblock


\bibitem[Qian et~al\mbox{.}(2025)]%
        {qian2025toolrl}
\bibfield{author}{\bibinfo{person}{Cheng Qian}, \bibinfo{person}{Emre~Can
  Acikgoz}, \bibinfo{person}{Qi He}, \bibinfo{person}{Hongru Wang},
  \bibinfo{person}{Xiusi Chen}, \bibinfo{person}{Dilek Hakkani-Tur},
  \bibinfo{person}{Gokhan Tur}, {and} \bibinfo{person}{Heng Ji}.}
  \bibinfo{year}{2025}\natexlab{}.
\newblock \showarticletitle{{ToolRL}: Reward is All Tool Learning Needs}. In
  \bibinfo{booktitle}{\emph{Advances in Neural Information Processing
  Systems}}, Vol.~\bibinfo{volume}{38}.
\newblock


\bibitem[Ren et~al\mbox{.}(2019)]%
        {ren2019time}
\bibfield{author}{\bibinfo{person}{Hansheng Ren}, \bibinfo{person}{Bixiong Xu},
  \bibinfo{person}{Yaoliang Wang}, \bibinfo{person}{Qi Yi}, {et~al\mbox{.}}}
  \bibinfo{year}{2019}\natexlab{}.
\newblock \showarticletitle{Time-Series Anomaly Detection Service at
  Microsoft}. In \bibinfo{booktitle}{\emph{Proceedings of the 25th ACM SIGKDD
  International Conference on Knowledge Discovery \& Data Mining}}.
  \bibinfo{pages}{3009--3017}.
\newblock


\bibitem[Schick et~al\mbox{.}(2023)]%
        {schick2023toolformer}
\bibfield{author}{\bibinfo{person}{Timo Schick}, \bibinfo{person}{Jane
  Dwivedi-Yu}, \bibinfo{person}{Roberto Dessi}, \bibinfo{person}{Roberta
  Raileanu}, \bibinfo{person}{Maria Lomeli}, \bibinfo{person}{Eric Hambro},
  \bibinfo{person}{Luke Zettlemoyer}, \bibinfo{person}{Nicola Cancedda}, {and}
  \bibinfo{person}{Thomas Scialom}.} \bibinfo{year}{2023}\natexlab{}.
\newblock \showarticletitle{{Toolformer}: Language Models Can Teach Themselves
  to Use Tools}. In \bibinfo{booktitle}{\emph{Advances in Neural Information
  Processing Systems}}, Vol.~\bibinfo{volume}{36}.
\newblock


\bibitem[Schmidl et~al\mbox{.}(2022)]%
        {schmidl2022anomaly}
\bibfield{author}{\bibinfo{person}{Sebastian Schmidl}, \bibinfo{person}{Phillip
  Wenig}, {and} \bibinfo{person}{Thorsten Papenbrock}.}
  \bibinfo{year}{2022}\natexlab{}.
\newblock \showarticletitle{Anomaly detection in time series: a comprehensive
  evaluation}.
\newblock \bibinfo{journal}{\emph{Proceedings of the VLDB Endowment}}
  \bibinfo{volume}{15}, \bibinfo{number}{9} (\bibinfo{year}{2022}),
  \bibinfo{pages}{1779--1797}.
\newblock


\bibitem[Shinn et~al\mbox{.}(2023)]%
        {shinn2023reflexion}
\bibfield{author}{\bibinfo{person}{Noah Shinn}, \bibinfo{person}{Federico
  Cassano}, \bibinfo{person}{Ashwin Gopinath}, \bibinfo{person}{Karthik
  Narasimhan}, {and} \bibinfo{person}{Shunyu Yao}.}
  \bibinfo{year}{2023}\natexlab{}.
\newblock \showarticletitle{{Reflexion}: Language Agents with Verbal
  Reinforcement Learning}. In \bibinfo{booktitle}{\emph{Advances in Neural
  Information Processing Systems}}, Vol.~\bibinfo{volume}{36}.
\newblock


\bibitem[Shyu et~al\mbox{.}(2003)]%
        {shyu2003novel}
\bibfield{author}{\bibinfo{person}{Mei-Ling Shyu}, \bibinfo{person}{Shu-Ching
  Chen}, \bibinfo{person}{Kanoksri Sarinnapakorn}, {and} \bibinfo{person}{LiWu
  Chang}.} \bibinfo{year}{2003}\natexlab{}.
\newblock \showarticletitle{A novel anomaly detection scheme based on principal
  component classifier}. In \bibinfo{booktitle}{\emph{Proceedings of the IEEE
  Foundations and New Directions of Data \mbox{Mining Workshop}}}.
  \bibinfo{pages}{172--179}.
\newblock


\bibitem[Su et~al\mbox{.}(2019)]%
        {su2019robust}
\bibfield{author}{\bibinfo{person}{Ya Su}, \bibinfo{person}{Youjian Zhao},
  \bibinfo{person}{Chenhao Niu}, \bibinfo{person}{Rong Liu},
  \bibinfo{person}{Wei Sun}, {and} \bibinfo{person}{Dan Pei}.}
  \bibinfo{year}{2019}\natexlab{}.
\newblock \showarticletitle{Robust anomaly detection for multivariate time
  series through stochastic recurrent neural network}. In
  \bibinfo{booktitle}{\emph{Proceedings of the 25th ACM SIGKDD International
  Conference on Knowledge Discovery \& Data Mining}}.
  \bibinfo{pages}{2828--2837}.
\newblock


\bibitem[Tang et~al\mbox{.}(2024)]%
        {tang2024mlp}
\bibfield{author}{\bibinfo{person}{Qideng Tang}, \bibinfo{person}{Chaofan Dai},
  \bibinfo{person}{Yahui Wu}, {and} \bibinfo{person}{Haohao Zhou}.}
  \bibinfo{year}{2024}\natexlab{}.
\newblock \showarticletitle{Mlp-mixer based masked autoencoders are effective,
  explainable and robust for time series anomaly detection}.
\newblock \bibinfo{journal}{\emph{Proceedings of the VLDB Endowment}}
  \bibinfo{volume}{18}, \bibinfo{number}{3} (\bibinfo{year}{2024}),
  \bibinfo{pages}{798--811}.
\newblock


\bibitem[Thill et~al\mbox{.}(2017)]%
        {thill2017time}
\bibfield{author}{\bibinfo{person}{Markus Thill}, \bibinfo{person}{Wolfgang
  Konen}, {and} \bibinfo{person}{Thomas B{\"a}ck}.}
  \bibinfo{year}{2017}\natexlab{}.
\newblock \showarticletitle{Time series anomaly detection with discrete wavelet
  transforms and maximum likelihood estimation}. In
  \bibinfo{booktitle}{\emph{Proceedings of the International Conference on Time
  Series and Forecasting}}, Vol.~\bibinfo{volume}{2}. \bibinfo{pages}{11--23}.
\newblock


\bibitem[Tuli et~al\mbox{.}(2022)]%
        {tuli2022tranad}
\bibfield{author}{\bibinfo{person}{Shreshth Tuli}, \bibinfo{person}{Giuliano
  Casale}, {and} \bibinfo{person}{Nicholas~R Jennings}.}
  \bibinfo{year}{2022}\natexlab{}.
\newblock \showarticletitle{TranAD: deep transformer networks for anomaly
  detection in multivariate time series data}.
\newblock \bibinfo{journal}{\emph{Proceedings of the VLDB Endowment}}
  \bibinfo{volume}{15}, \bibinfo{number}{6} (\bibinfo{year}{2022}),
  \bibinfo{pages}{1201--1214}.
\newblock


\bibitem[Wang et~al\mbox{.}(2024)]%
        {wang2024fcvae}
\bibfield{author}{\bibinfo{person}{Zexin Wang}, \bibinfo{person}{Changhua Pei},
  \bibinfo{person}{Minghua Ma}, \bibinfo{person}{Xin Wang},
  \bibinfo{person}{Zhihan Li}, \bibinfo{person}{Dan Pei},
  \bibinfo{person}{Saravan Rajmohan}, \bibinfo{person}{Dongmei Zhang},
  \bibinfo{person}{Qingwei Lin}, \bibinfo{person}{Haiming Zhang},
  \bibinfo{person}{Jianhui Li}, {and} \bibinfo{person}{Gaogang Xie}.}
  \bibinfo{year}{2024}\natexlab{}.
\newblock \showarticletitle{Revisiting {VAE} for Unsupervised Time Series
  Anomaly Detection: A Frequency Perspective}. In
  \bibinfo{booktitle}{\emph{Proceedings of the ACM on Web Conference 2024}}.
  \bibinfo{publisher}{ACM}, \bibinfo{address}{New York, NY, USA},
  \bibinfo{pages}{3096--3105}.
\newblock


\bibitem[Wu et~al\mbox{.}(2026)]%
        {wu2026icad}
\bibfield{author}{\bibinfo{person}{Zhongyuan Wu}, \bibinfo{person}{Jingyuan
  Wang}, \bibinfo{person}{Zexuan Cheng}, \bibinfo{person}{Yilong Zhou},
  \bibinfo{person}{Weizhi Wang}, \bibinfo{person}{Juhua Pu},
  \bibinfo{person}{Chao Li}, {and} \bibinfo{person}{Changqing Ma}.}
  \bibinfo{year}{2026}\natexlab{}.
\newblock \showarticletitle{{ICAD-LLM}: One-for-All Anomaly Detection via
  In-Context Learning with Large Language Models}. In
  \bibinfo{booktitle}{\emph{Proceedings of the AAAI Conference on Artificial
  Intelligence}}, Vol.~\bibinfo{volume}{40}. \bibinfo{pages}{15986--15994}.
\newblock


\bibitem[Xu et~al\mbox{.}(2018)]%
        {xu2018unsupervised}
\bibfield{author}{\bibinfo{person}{Haowen Xu}, \bibinfo{person}{Wenxiao Chen},
  \bibinfo{person}{Nengwen Zhao}, \bibinfo{person}{Zeyan Li},
  \bibinfo{person}{Jiahao Bu}, \bibinfo{person}{Zhihan Li},
  \bibinfo{person}{Ying Liu}, \bibinfo{person}{Youjian Zhao},
  \bibinfo{person}{Dan Pei}, \bibinfo{person}{Yang Feng}, {et~al\mbox{.}}}
  \bibinfo{year}{2018}\natexlab{}.
\newblock \showarticletitle{Unsupervised anomaly detection via variational
  auto-encoder for seasonal kpis in web applications}. In
  \bibinfo{booktitle}{\emph{Proceedings of the 2018 World Wide Web
  Conference}}. \bibinfo{pages}{187--196}.
\newblock


\bibitem[Xu et~al\mbox{.}(2022)]%
        {xuanomaly}
\bibfield{author}{\bibinfo{person}{Jiehui Xu}, \bibinfo{person}{Haixu Wu},
  \bibinfo{person}{Jianmin Wang}, {and} \bibinfo{person}{Mingsheng Long}.}
  \bibinfo{year}{2022}\natexlab{}.
\newblock \showarticletitle{Anomaly Transformer: Time Series Anomaly Detection
  with Association Discrepancy}. In \bibinfo{booktitle}{\emph{International
  Conference on Learning Representations}}.
\newblock


\bibitem[Xu et~al\mbox{.}(2026)]%
        {xu2026multimodal}
\bibfield{author}{\bibinfo{person}{Xiongxiao Xu}, \bibinfo{person}{Haoran
  Wang}, \bibinfo{person}{Yueqing Liang}, \bibinfo{person}{Philip~S. Yu},
  \bibinfo{person}{Yue Zhao}, {and} \bibinfo{person}{Kai Shu}.}
  \bibinfo{year}{2026}\natexlab{}.
\newblock \showarticletitle{Can Multimodal {LLMs} Perform Time Series Anomaly
  Detection?}. In \bibinfo{booktitle}{\emph{Proceedings of the ACM Web
  Conference 2026}}. \bibinfo{publisher}{ACM}, \bibinfo{address}{New York, NY,
  USA}, \bibinfo{pages}{5392--5403}.
\newblock


\bibitem[Yang et~al\mbox{.}(2025)]%
        {yang2025qwen3}
\bibfield{author}{\bibinfo{person}{An Yang}, \bibinfo{person}{Anfeng Li},
  \bibinfo{person}{Baosong Yang}, \bibinfo{person}{Beichen Zhang},
  \bibinfo{person}{Binyuan Hui}, \bibinfo{person}{Bo Zheng},
  \bibinfo{person}{Bowen Yu}, \bibinfo{person}{Chang Gao},
  \bibinfo{person}{Chengen Huang}, \bibinfo{person}{Chenxu Lv},
  {et~al\mbox{.}}} \bibinfo{year}{2025}\natexlab{}.
\newblock \bibinfo{title}{Qwen3 Technical Report}.
\newblock
\shownote{arXiv:2505.09388}.
\newblock


\bibitem[Yang et~al\mbox{.}(2026)]%
        {yang2026timera}
\bibfield{author}{\bibinfo{person}{Yiyuan Yang}, \bibinfo{person}{Zichuan Liu},
  \bibinfo{person}{Lei Song}, \bibinfo{person}{Kai Ying},
  \bibinfo{person}{Stephen Wang}, \bibinfo{person}{Joshua~Thomas Bamford},
  \bibinfo{person}{Svitlana Vyetrenko}, \bibinfo{person}{Jiang Bian}, {and}
  \bibinfo{person}{Qingsong Wen}.} \bibinfo{year}{2026}\natexlab{}.
\newblock \showarticletitle{{Time-RA}: Towards Time Series Reasoning for
  Anomaly Diagnosis with {LLM} Feedback}. In \bibinfo{booktitle}{\emph{Findings
  of the Association for Computational Linguistics: ACL 2026}}.
  \bibinfo{publisher}{Association for Computational Linguistics},
  \bibinfo{pages}{11591--11616}.
\newblock


\bibitem[Yao et~al\mbox{.}(2023)]%
        {yao2023react}
\bibfield{author}{\bibinfo{person}{Shunyu Yao}, \bibinfo{person}{Jeffrey Zhao},
  \bibinfo{person}{Dian Yu}, \bibinfo{person}{Nan Du}, \bibinfo{person}{Izhak
  Shafran}, \bibinfo{person}{Karthik Narasimhan}, {and} \bibinfo{person}{Yuan
  Cao}.} \bibinfo{year}{2023}\natexlab{}.
\newblock \showarticletitle{{ReAct}: Synergizing Reasoning and Acting in
  Language Models}. In \bibinfo{booktitle}{\emph{International Conference on
  Learning Representations}}.
\newblock


\bibitem[Yeh et~al\mbox{.}(2016)]%
        {yeh2016matrix}
\bibfield{author}{\bibinfo{person}{Chin-Chia~Michael Yeh}, \bibinfo{person}{Yan
  Zhu}, \bibinfo{person}{Liudmila Ulanova}, \bibinfo{person}{Nurjahan Begum},
  \bibinfo{person}{Yifei Ding}, \bibinfo{person}{Hoang~Anh Dau},
  \bibinfo{person}{Diego~Furtado Silva}, \bibinfo{person}{Abdullah Mueen},
  {and} \bibinfo{person}{Eamonn Keogh}.} \bibinfo{year}{2016}\natexlab{}.
\newblock \showarticletitle{Matrix profile I: All pairs similarity joins for
  time series: A unifying view that includes motifs, discords and shapelets}.
  In \bibinfo{booktitle}{\emph{2016 IEEE 16th International Conference on Data
  Mining}}. IEEE, \bibinfo{pages}{1317--1322}.
\newblock


\bibitem[Zhang et~al\mbox{.}(2017)]%
        {zhang2017cleaning}
\bibfield{author}{\bibinfo{person}{Aoqian Zhang}, \bibinfo{person}{Shaoxu
  Song}, \bibinfo{person}{Jianmin Wang}, {and} \bibinfo{person}{Philip~S. Yu}.}
  \bibinfo{year}{2017}\natexlab{}.
\newblock \showarticletitle{Time Series Data Cleaning: From Anomaly Detection
  to Anomaly Repairing}.
\newblock \bibinfo{journal}{\emph{Proceedings of the VLDB Endowment}}
  \bibinfo{volume}{10}, \bibinfo{number}{10} (\bibinfo{year}{2017}),
  \bibinfo{pages}{1046--1057}.
\newblock


\bibitem[Zhang et~al\mbox{.}(2019)]%
        {zhang2019mscred}
\bibfield{author}{\bibinfo{person}{Chuxu Zhang}, \bibinfo{person}{Dongjin
  Song}, \bibinfo{person}{Yuncong Chen}, \bibinfo{person}{Xinyang Feng},
  \bibinfo{person}{Cristian Lumezanu}, \bibinfo{person}{Wei Cheng},
  \bibinfo{person}{Jingchao Ni}, \bibinfo{person}{Bo Zong},
  \bibinfo{person}{Haifeng Chen}, {and} \bibinfo{person}{Nitesh~V. Chawla}.}
  \bibinfo{year}{2019}\natexlab{}.
\newblock \showarticletitle{A Deep Neural Network for Unsupervised Anomaly
  Detection and Diagnosis in Multivariate Time Series Data}. In
  \bibinfo{booktitle}{\emph{Proceedings of the AAAI Conference on Artificial
  Intelligence}}, Vol.~\bibinfo{volume}{33}. \bibinfo{pages}{1409--1416}.
\newblock


\bibitem[Zhang et~al\mbox{.}(2026b)]%
        {zhang2026anomseer}
\bibfield{author}{\bibinfo{person}{Junru Zhang}, \bibinfo{person}{Lang Feng},
  \bibinfo{person}{Haoran Shi}, \bibinfo{person}{Xu Guo}, \bibinfo{person}{Han
  Yu}, \bibinfo{person}{Yabo Dong}, {and} \bibinfo{person}{Duanqing Xu}.}
  \bibinfo{year}{2026}\natexlab{b}.
\newblock \showarticletitle{{AnomSeer}: Reinforcing Multimodal {LLMs} to Reason
  for Time-Series Anomaly Detection}. In \bibinfo{booktitle}{\emph{Proceedings
  of the 43rd International Conference on Machine Learning}}.
  \bibinfo{pages}{155426--155449}.
\newblock


\bibitem[Zhang et~al\mbox{.}(2022)]%
        {zhang2022efficient}
\bibfield{author}{\bibinfo{person}{Shenglin Zhang}, \bibinfo{person}{Zhenyu
  Zhong}, \bibinfo{person}{Dongwen Li}, \bibinfo{person}{Qiliang Fan},
  \bibinfo{person}{Yongqian Sun}, \bibinfo{person}{Man Zhu},
  \bibinfo{person}{Yuzhi Zhang}, \bibinfo{person}{Dan Pei},
  \bibinfo{person}{Jiyan Sun}, \bibinfo{person}{Yinlong Liu}, {et~al\mbox{.}}}
  \bibinfo{year}{2022}\natexlab{}.
\newblock \showarticletitle{Efficient kpi anomaly detection through transfer
  learning for large-scale web services}.
\newblock \bibinfo{journal}{\emph{IEEE Journal on Selected Areas in
  Communications}} \bibinfo{volume}{40}, \bibinfo{number}{8}
  (\bibinfo{year}{2022}), \bibinfo{pages}{2440--2455}.
\newblock


\bibitem[Zhang et~al\mbox{.}(2026a)]%
        {zhang2026search}
\bibfield{author}{\bibinfo{person}{Wenlin Zhang}, \bibinfo{person}{Kuicai
  Dong}, \bibinfo{person}{Junyi Li}, \bibinfo{person}{Yingyi Zhang},
  \bibinfo{person}{Xiaopeng Li}, \bibinfo{person}{Pengyue Jia},
  \bibinfo{person}{Yi Wen}, \bibinfo{person}{Derong Xu},
  \bibinfo{person}{Maolin Wang}, \bibinfo{person}{Yichao Wang},
  \bibinfo{person}{Yong Liu}, {and} \bibinfo{person}{Xiangyu Zhao}.}
  \bibinfo{year}{2026}\natexlab{a}.
\newblock \showarticletitle{To Search or Not to Search: Aligning the Decision
  Boundary of Deep Search Agents via Causal Intervention}. In
  \bibinfo{booktitle}{\emph{Proceedings of the ACM Web Conference 2026}}.
  \bibinfo{publisher}{ACM}, \bibinfo{address}{New York, NY, USA},
  \bibinfo{pages}{2049--2059}.
\newblock


\bibitem[Zhou et~al\mbox{.}(2023)]%
        {zhou2023one}
\bibfield{author}{\bibinfo{person}{Tian Zhou}, \bibinfo{person}{Peisong Niu},
  \bibinfo{person}{Liang Sun}, \bibinfo{person}{Rong Jin}, {et~al\mbox{.}}}
  \bibinfo{year}{2023}\natexlab{}.
\newblock \showarticletitle{One fits all: Power general time series analysis by
  pretrained lm}.
\newblock \bibinfo{journal}{\emph{Advances in Neural Information Processing
  Systems}}  \bibinfo{volume}{36} (\bibinfo{year}{2023}),
  \bibinfo{pages}{43322--43355}.
\newblock


\end{thebibliography}

\appendix
\section{Implementation and Training}
\label{app:method-implementation}
\subsection{Knowledge and Numerical Tools}
\label{app:knowledge-tools}
As described in Section~\ref{sec:adaptive-feature-preparation}, the static knowledge memory $K$ supplies role-specific prompt contexts $K_{loc}$, $K_{act}$, and $K_{det}$ rather than a callable retrieval tool. Its dataset-specific guidance on anomaly patterns, boundaries, thresholds, and tool use is derived from training data and public domain knowledge and fixed before evaluation, without test annotations or results. YAHOO favors tight spike or dip intervals excluding recovery points; KPI, IOPS, and WSD consider neighboring observations around isolated spikes and complete shift--recovery intervals. Suggested value/change z-score threshold pairs are $(3,3)$, $(3\text{--}3.5,4)$, $(3.5,4)$, and $(3.5,5)$ for YAHOO, KPI, IOPS, and WSD, respectively. These are diagnostic priors, not enforced decision rules.

The locator receives an unannotated time-series plot. Numerical references $B$ contain the means and standard deviations of values and first-order differences, computed from all preprocessed observations $G$ in the complete parent training or test split before window construction. These references can include observations beyond the current window, as specified by the evaluation protocol in Section~\ref{sec:experimental-settings} and Appendix~\ref{app:data-preprocessing}; test anomaly labels are not used. Local statistics are computed over the current segment. The diagnostic engines $N_{stat}$, $N_{val}$, $N_{chg}$, and $N_{reg}$ correspond to \texttt{stats}, \texttt{value\_zscore}, \texttt{diff\_zscore}, and \texttt{grep}, respectively. Value-based and change-based diagnostics compute standardized deviations as:

\[
z_t^{val}=\frac{x_t-\mu_G}{\sigma_G},\qquad
z_t^{chg}=\frac{\Delta x_t-\mu_{\Delta,G}}{\sigma_{\Delta,G}},
\]

where $\Delta x_t=x_t-x_{t-1}$ is assigned to the later observation. The actor provides inspection thresholds returning reference statistics and indices grouped by deviation magnitude. Sorted thresholds partition observations into $\tau_i<|z_t|\leq\tau_{i+1}$ and $|z_t|>\tau_m$. \texttt{grep(start, end)} retrieves inclusive in-range index--value pairs. Nonpositive standard deviations are replaced with $10^{-12}$, while missing references or insufficient observations yield structured errors.

\subsection{Role Protocols, Parsing, and Control}
\label{app:workflow-control}
Inclusive intervals use split-local indices: window positions $1,\ldots,T$ map to $s,\ldots,s+T-1$. Structured outputs contain no Markdown fences.
The locator identifies suspicious regions as candidate intervals $C$ from the chart and visual-domain rules. The actor uses these candidates, domain rules, tool schemas, and review feedback to batch diagnostics and refine regional queries. Regional spans of at most 200 points are advised rather than enforced.
The detector receives $C$, actor context and the current-pass transcript $Z^{(r)}$, domain guidance $K_{det}$, and latest feedback $f^{(r)}$, forming $\mathbf d^{(r)}$ in Section~\ref{sec:reasoning-aware-detection}. KPI guidance includes evidence-priority rules. The evaluator reviews one-sentence summaries of detector results, actor evidence/tool use, and prior guidance, without full transcripts, separate knowledge, or test labels. It requests redos for serious inconsistencies or missed anomalies and accepts minor or uncertain cases. Appendix~\ref{app:prompt-templates} specifies each role's structured output.

\label{sec:refinement-control}

The inference parser scans after closed reasoning before checking the full response, allowing optional reasoning and trailing text. It skips invalid entries, converts endpoints to integers, reverses unordered intervals, and treats single endpoints as points. Confidence defaults to 2 when missing or invalid, with values $\geq2$ mapped to 2 and others to 1. Only observed covered indices are scored. Unparseable reviews imply acceptance.
A redo clears actor/detector histories, transcript $Z^{(r)}$, and outputs while retaining candidates $C$ and updated feedback $f^{(r+1)}$. Each pass allows six actor turns with multiple tool calls, resetting on redo. The zero-based condition $>3$ allows four redos and five detector calls. Outer recovery permits three attempts by default.

\subsection{Training and Inference Details}
\label{app:training-inference}

\paragraph{Training data construction.}
In the default hybrid configuration described in Section~\ref{sec:experimental-settings}, Qwen3-8B serves as the trainable detector, while the locator, actor, and evaluator use frozen Gemini-3.1-flash. SFT data are constructed from Gemini-3.1-flash diagnostic trajectories, pairing detector prompts reconstructed from accumulated evidence with recorded detector responses, rather than answers derived from ground-truth intervals. Empty samples and prompts exceeding 7,000 tokens are excluded, without performance-based filtering. Both normal and anomalous windows are included, with missing confidence values defaulting to 2. Ground-truth intervals are used only for RL reward computation.

\paragraph{Training configuration.}
In the default hybrid configuration, SFT and RL update all detector parameters without LoRA adaptation. SFT uses the Transformers Trainer with a learning rate of $5\times10^{-5}$, a per-device batch size of 2, gradient accumulation of 4, and one training epoch. We adopt a constant learning rate schedule without warmup or weight decay. Chat serialization appends an EOS token before truncation to a maximum combined length of 9,000 tokens, with prompt and padding tokens masked from the training loss.
RL training is conducted on eight Ascend NPUs using GRPO. The learning rate is $10^{-6}$, with a training batch size of 32, a per-device micro-batch size of 8, and two data epochs. Each update uses one PPO epoch. The group size is $G=8$, with sampling temperature 1, top-$p=0.95$, and top-$k=150$. The maximum prompt, response, and model sequence lengths are 3,000, 1,024, and 4,025 tokens, respectively. Overlong prompts are excluded. The optimization uses neither a separate critic nor KL regularization. Data preparation uses random seed 42, while rollout seeds follow replica ranks.

\paragraph{Validation and inference.}
Training follows fixed hyperparameter settings and predefined epoch schedules, with the final checkpoint used for inference. Test annotations are reserved for evaluation and are not used for hyperparameter or checkpoint selection. The default sampling temperature is 0.6 for all inference roles; Section~\ref{sec:temperature-analysis} separately varies the detector temperature. Results are averaged over three inference runs using the same trained checkpoint.

\subsection{Learning Objectives}
\label{app:learning-objectives}

\paragraph{Detector reward.}
\label{sec:detector-reward}
In the default hybrid configuration, SFT follows Section~\ref{sec:detection-policy-optimization}, optimizing only the detector. During RL, let $P$ be the integer time points covered by intervals parsed from response $u$, and let $Y$ be the anomalous points in training labels $\mathbf y$. Overlapping intervals are deduplicated. The reward is:
\[
R(u,\mathbf y)=
\begin{cases}
-1, & \text{invalid response format},\\
0, & \text{trailing text after the JSON list},\\
\frac{2|P\cap Y|}{|P|+|Y|}, & |Y|>0,\\
\frac{1}{|P|+1}, & |Y|=0.
\end{cases}
\]
A valid training response requires a closed \texttt{think}, \texttt{redacted\_}\allowbreak\texttt{reasoning}, or \texttt{redacted\_}\allowbreak\texttt{thinking} block followed by a JSON list. These reward-format checks are stricter than the inference parser in Appendix~\ref{app:workflow-control}. Single-point and reversed endpoints are accepted; explanations, confidence, and interval bounds receive no additional validation. Unusable entries are excluded. No parsing bonus or sample weighting is applied.

\paragraph{GRPO updates.}
\label{sec:grpo-updates}
All $G=8$ sampled detector responses within an RL group share the same assembled diagnostic context $\mathbf d$: candidates, tool evidence, domain guidance, and review feedback are held fixed. In the default hybrid configuration, both SFT and RL update only the detector; the locator, actor, and evaluator remain frozen. Following Section~\ref{sec:detection-policy-optimization}, rewards are normalized using the sample standard deviation with denominator $G-1$ and a stabilizer of $10^{-6}$. Groups with identical rewards yield zero advantages. The resulting advantage $\hat A_i$ is shared across response tokens within each sampled completion.
For diagnostic context $\mathbf d$ and preceding response tokens $u_{i,<t}$, let $\delta_{i,t}$ denote the difference between current and old token log probabilities.
The optimization objective is:
\begin{equation}
\begin{aligned}
\rho_{i,t}
&=\exp\{\operatorname{clip}(\delta_{i,t},-20,20)\},\\
b_{i,t}
&=\min\{\rho_{i,t}\hat A_i,
\operatorname{clip}(\rho_{i,t},0.8,1.28)\hat A_i\},\\
q_{i,t}
&=\begin{cases}
b_{i,t},&\hat A_i\geq0,\\
\max\{b_{i,t},10\hat A_i\},&\hat A_i<0,
\end{cases}\\
L_{RL}
&=-\frac{\sum_{i,t}m_{i,t}q_{i,t}}
{\sum_{i,t}m_{i,t}}.
\end{aligned}
\end{equation}
Here, $m_{i,t}$ masks prompt and padding tokens, and summation is performed over the update batch. The objective does not include a KL penalty, and surrogate clipping does not impose a strict bound on every realized probability ratio.

\section{Experimental Details}
\label{app:experimental-protocol}

\subsection{Data Construction and Preprocessing}
\label{app:datasets}
\label{app:sample-construction}

Table~\ref{tab:dataset} summarizes coverage. Tested sequences use $\lfloor T/2\rfloor$ initial observations for training and the rest for testing; training-only sequences use all observations. Splits are independently preprocessed before windowing. For YAHOO, KPI, IOPS, and WSD, window lengths are 100, 1,000, 1,000, and 500, with 1,072, 483, 147, and 178 training windows, respectively. Repetitions of 1, 2, 3, and 3 yield 3,013 multi-domain instances from 1,880 distinct windows per stage. Strides equal window lengths; incomplete tails are discarded, and windows never cross sequence or split boundaries. Testing retains all complete windows without repetition or label-based subsampling; sequence eligibility applies only during metric aggregation.

\paragraph{Preprocessing.}
\label{app:data-preprocessing}
Each complete training or test split is preprocessed independently before segmentation. Nonfinite observations are treated as missing values and filled through bidirectional linear interpolation, followed by boundary filling and median imputation when necessary; an entirely missing split is filled with zeros. Anomaly labels remain unchanged.

\emph{Normalization.} Following prior work~\cite{su2019robust}, we normalize each complete training or test split independently before window construction. Let $a$ and $b$ be the minimum and maximum values of the complete parent split. For $b>a$, min--max scaling maps observations to $[0,1]$:
\[
u_i=\frac{x_i-a}{b-a}.
\]
Constant splits are mapped to zero. \emph{Indexing.} Original timestamps are replaced with zero-based integer indices $i=0,\ldots,N-1$, where $N$ is the split length; these indices are preserved during window construction. \emph{Detrending.} We fit a linear trend $\hat{u}_{\mathrm{trend},i}=\hat{\alpha}i+\hat{\beta}$ to the normalized split by least squares and subtract it, yielding $\tilde{x}_i=u_i-\hat{u}_{\mathrm{trend},i}$. The resulting values are rounded to six decimal places and need not remain in $[0,1]$.
All preprocessing statistics are computed from the complete unlabeled split before window construction. Test anomaly labels are used exclusively for evaluation and are not involved in preprocessing or diagnostic decisions.

\subsection{Baseline Implementations}
\label{app:baselines}

All methods use identical test windows and splits as specified in Section~\ref{sec:experimental-settings} and share the preprocessing in Appendix~\ref{app:data-preprocessing}: missing-value imputation, min--max scaling, zero-based indexing, and linear detrending before window construction. Supervised NCAD and AnomaMind-Direct share AnomaMind's labeled training splits. AnomaMind-Direct uses SFT to train Qwen3-8B for interval prediction directly from input windows, without diagnostic interactions. AnomaMind-Zero retains the complete workflow with frozen Gemini-3.1-flash in all four roles. In contrast, the \textit{w/o both} training ablation retains the default role assignments but applies neither SFT nor RL to its Qwen3-8B detector; \textit{w/o RL} retains SFT and the complete workflow. These comparisons differ in context, model assignments, or training and computational costs.

Figure~\ref{fig:training} compares detector-only, actor-only, and actor plus detector training. Each trained role uses Qwen3-8B with SFT followed by RL, while other roles remain frozen general-purpose LLMs. Actor-only trains the actor with a frozen general-purpose detector, whereas actor plus detector trains both roles. Actor RL uses the same detection reward as detector RL, computed from the final detector response. This comparison examines whether jointly optimizing roles offers additional benefits.

\paragraph{Tool-use strategies.}
\label{app:tool-use-strategies}
Fixed, random, and adaptive share cached locator outputs, detector checkpoint, domain knowledge, evaluator configuration, and other inference settings. Fixed invokes \texttt{stats}, \texttt{value\_zscore}, and \texttt{diff\_zscore} per window, with value/change thresholds of $(3,3)$, $(3.5,4)$, $(3.5,5)$, and $(3.5,4)$ for YAHOO, KPI, WSD, and IOPS, respectively. Region inspection via \texttt{grep} runs only for locator candidates; \texttt{stats} and \texttt{grep} require no thresholds. Per window, random samples per-tool invocation probabilities $p\sim\mathcal{U}(0,1)$, requiring at least one invocation. Selected value/change thresholds follow $\mathcal{N}(\mu,(0.1\mu)^2)$ truncated to $[0.8\mu,1.2\mu]$, with $\mu$ denoting the corresponding fixed threshold. Fixed and random inspect locator candidates, merging overlapping intervals and querying disjoint intervals separately. Without candidates, random invokes no tools and fixed skips only region inspection. Adaptive determines tools, thresholds, and query regions from accumulated evidence, allowing numerical signals to trigger inspection without visual candidates and further queries of neighboring points. Strategy sampling uses seed 17; we average results over three inference runs. Tool use counts individual executions per test window, including zero-call windows; calls count separately.

\subsection{Metrics and Aggregation}
\label{app:evaluation-metrics}
The evaluation script groups complete test windows by parent sequence, orders them by segment ID, and concatenates their predictions and labels. Consistent with the anomaly-containing evaluation setting in TSAD-Agents~\cite{xu2026multimodal}, we exclude anomaly-free sequences from macro-averaging but retain all complete windows of eligible sequences, including normal windows; false alarms on normal points are counted as false positives (FP). Thus, false alarms on entirely normal sequences are not evaluated. Records with \texttt{final\_output.error}, unreadable results, missing segments, or invalid labels are excluded. All reported AnomaMind runs achieve 100\% coverage of constructed test windows. Sequence-level metrics are macro-averaged over the same eligible sequences for all methods, with results averaged over three inference runs.

Following prior TSAD research~\cite{tang2024mlp}, continuous-score baselines use a fixed threshold of $\mu_{\mathrm{train}}+3\sigma_{\mathrm{train}}$, where the statistics are estimated from training anomaly scores. This protocol avoids test-label-based threshold calibration, although it may differ from method-specific defaults and affect threshold-dependent metrics. We therefore additionally report AUC-PR and oracle best-F1 to provide complementary evaluations.
Point-wise F1 uses \texttt{zero\_division=0}, computed per sequence rather than from averaged precision and recall. False positives on normal points, including those in entirely normal windows of eligible sequences, contribute to precision and F1. AnomaMind F1 and range-F1 use the same binary interval coverage. The confidence mapping defined in Appendix~\ref{app:workflow-control} gives covered points scores 1 or 2, overlaps their maximum, and uncovered points 0; these are ordinal scores, not probabilities. Best-F1 sweeps $\{0.5\}\cup\{s_t:s_t>0\}$ per sequence, using evaluation labels without changing ordinary predictions; continuous baselines likewise sweep their scores. Following prior TSAD research~\cite{paparrizos2022volume}, we report AUC-PR using the average precision (AP) formulation implemented by \texttt{average\_precision\_score}, which weights precision by recall increments rather than trapezoidal integration. AUC-PR and best-F1 are macro-averaged over finite values within the same eligible sequence set. Range-F1 uses TSB-AD R-based-F1~\cite{liu2024elephant},
overlap weight $\alpha=0.2$, on binary predictions without threshold search.

\subsection{Prompt Templates}
\label{app:prompt-templates}
These condensed templates summarize the four roles. Dataset rules, segment bounds, tool schemas/results, and retry guidance are injected at runtime. All indices are split-local; detailed protocols appear in Appendix~\ref{app:workflow-control}. Structured outputs use no Markdown fences.

\begin{StrategyBox}[frameblue]{bgblue}{Locator Role Template}
\refstepcounter{idx}
\label{prompt:planner}

\noindent\textbf{Input:} Segment chart, dataset type, and visual-domain rules.

\noindent\textbf{Task:} Read chart axes and propose a few broad intervals of abrupt change rather than slow drift. Merge nearby bursts from one event.

\noindent\textbf{Output:} A short summary enclosed by \texttt{<locator\_}\allowbreak\texttt{description>} and \texttt{</locator\_}\allowbreak\texttt{description>}, followed immediately by a JSON list of \texttt{interval} and \texttt{rationale} objects. If no candidate exists, explain this in the summary and return \texttt{[]}.
\end{StrategyBox}
\begin{StrategyBox}[frameblue]{bgblue}{Actor Role Template}
\refstepcounter{idx}
\label{prompt:executor}

\noindent\textbf{Input:} Locator output, bounds, domain rules, schemas, and feedback.

\noindent\textbf{Task:} Batch independent \texttt{stats}, \texttt{value\_zscore}, and \texttt{diff\_zscore} calls; add \texttt{grep(start,end)} for usable candidates. Adapt further queries to the evidence to confirm, reject, or refine candidates. Prefer regional queries of at most 200 points; broaden slightly for misalignment and avoid repeated queries.

\noindent\textbf{Output:} Summarize sufficient evidence in plain text.
\end{StrategyBox}

\begin{StrategyBox}[frameblue]{bgblue}{Detector Role Template}
\refstepcounter{idx}
\label{prompt:detector}

\noindent\textbf{Input:} Locator candidates, Actor context and current-pass query--result transcript, domain rules, and latest review feedback; KPI includes final-stage evidence-priority rules.

\noindent\textbf{Task:} Predict anomaly intervals from the accumulated evidence.

\noindent\textbf{Output:} \texttt{<think>...</think>} followed immediately by one JSON list with fields \texttt{interval: [start,end]}, \texttt{type}, \texttt{explanation}, and integer \texttt{confidence} set to 1 or 2. Use inclusive split-local endpoints. Return \texttt{[]} for no anomaly, and end at the list's closing bracket.
\end{StrategyBox}

\begin{StrategyBox}[frameblue]{bgblue}{Evaluator Role Template}
\refstepcounter{idx}
\label{prompt:checker}

\noindent\textbf{Input:} One-sentence summaries of Detector results, Actor evidence/tool use, and prior guidance; no transcripts or separate knowledge.

\noindent\textbf{Task:} Request a redo only for clearly unsupported intervals, obvious missed anomalies, or tool use that contradicts evidence or ignores critical regions. Accept minor, borderline, or uncertain cases.

\noindent\textbf{Output:} One JSON object with boolean \texttt{redo} and string \texttt{suggestion}. Set \texttt{redo} to \texttt{true} and provide actionable Actor guidance when revision is needed; otherwise return \texttt{\{"redo": false, "suggestion": ""\}}.
\end{StrategyBox}

\end{document}